\pdfoutput=1

\documentclass[11pt]{article}

\usepackage[final]{acl}

\usepackage{times}
\usepackage{latexsym}

\usepackage[T1]{fontenc}

\usepackage[utf8]{inputenc}

\usepackage{microtype}

\usepackage{inconsolata}

\usepackage{graphicx}

\usepackage{amsmath}
\usepackage{amssymb}
\usepackage{mathtools}
\usepackage{amsthm}
\usepackage{multirow}
\usepackage{bm}
\usepackage{xcolor}
\usepackage{colortbl}
\usepackage{booktabs}
\usepackage{chngcntr}
\usepackage{algorithm}
\usepackage{algpseudocode}

\usepackage{float}
\usepackage{graphicx}
\usepackage{wrapfig}
\usepackage{stfloats}
\usepackage{caption}
\usepackage{subcaption}
\usepackage{threeparttable}

\usepackage{listings}
\DeclarePairedDelimiter{\nint}\lfloor\rceil

\title{ClusComp: A Simple Paradigm for Model Compression \\ and Efficient Finetuning}

\author{
  Baohao Liao$^{1,2}$\thanks{Correspondence to: \texttt{\href{mailto:liaobaohao@gmail.com}{liaobaohao@gmail.com}}} \:\:\:\:\: Christian Herold$^{2}$ \:\:\:\:\:  Seyyed Hadi Hashemi$^{2}$ \\ \:\:\:\:\: \textbf{Stefan Vasilev}$^{1,2}$  \:\:\:\:\:  \textbf{Shahram Khadivi}$^{2}$ \:\:\:\:\: \textbf{Christof Monz}$^{1}$ \\
  $^{1}$Language Technology Lab, University of Amsterdam \\
  $^{2}$eBay Inc. \\
}

\begin{document}

\maketitle

\begin{figure*}[b]
  \centering
  \includegraphics[width=0.95\textwidth]{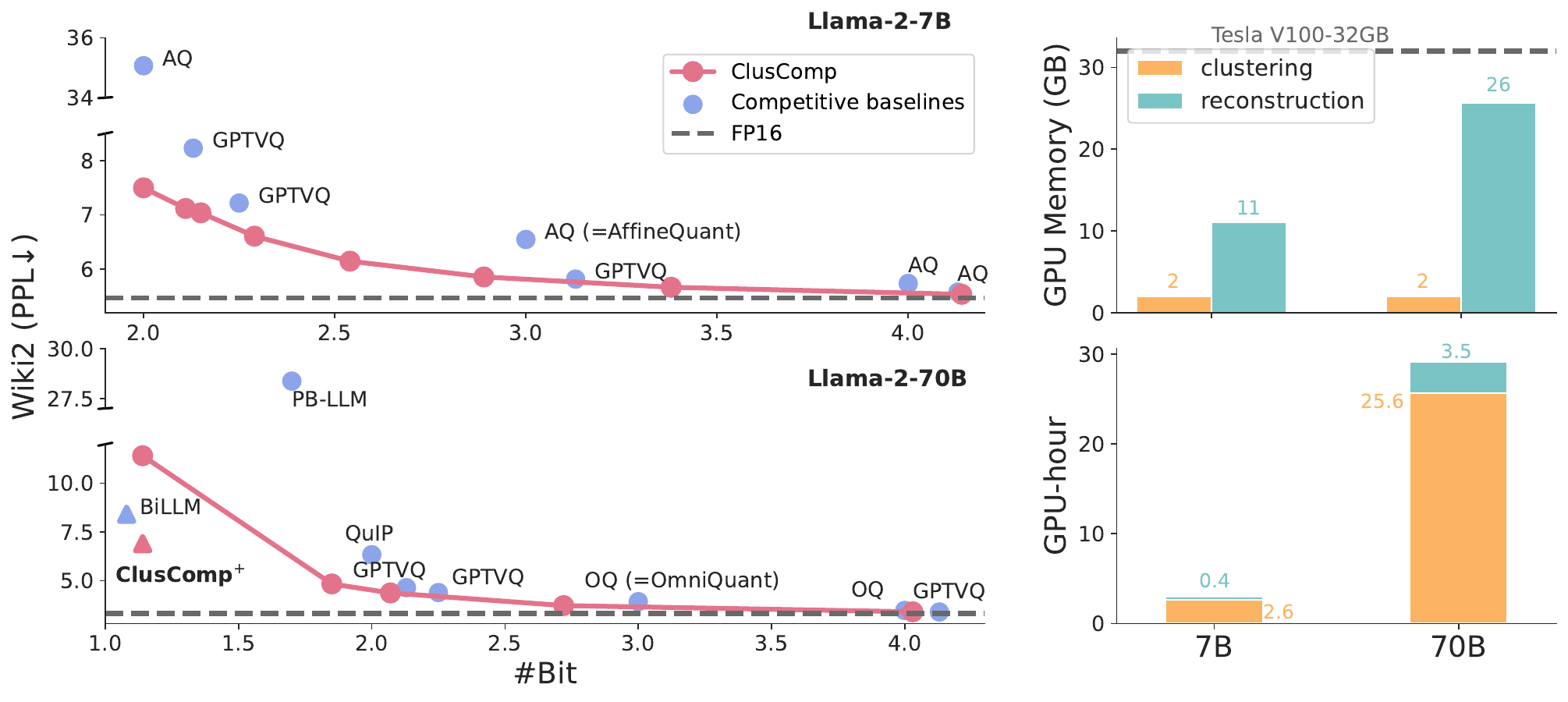}
  \caption[]{\textbf{Left:} Compression quality of ClusComp. Only the most competitive baselines are shown here. Please refer to Table \ref{tab: wiki ppl} and \ref{tab: c4 ppl} for a full comparison. Methods in triangle use more calibration samples than the ones in circle. \textbf{Right:} Compression efficiency of ClusComp that consists of sequential clustering and reconstruction steps. E.g., compressing a 70B LLM takes 25.6h + 2GB for clustering, and 3.5h + 26GB for reconstruction on 1 GPU.}
  \label{fig: overall}
\end{figure*}

\begin{abstract}
As large language models (LLMs) scale, model compression is crucial for edge deployment and accessibility. Weight-only quantization reduces model size but suffers from performance degradation at lower bit widths. Moreover, standard finetuning is incompatible with quantized models, and alternative methods often fall short of full finetuning. In this paper, we propose \textit{ClusComp}, a simple yet effective compression paradigm that clusters weight matrices into codebooks and finetunes them block-by-block. ClusComp (1) achieves superior performance in 2-4 bit quantization, (2) pushes compression to 1-bit while outperforming ultra-low-bit methods with minimal finetuning, and (3) enables efficient finetuning, even surpassing existing quantization-based approaches and rivaling full FP16 finetuning. Notably, ClusComp supports compression and finetuning of 70B LLMs on a single A6000-48GB GPU.
\end{abstract}

\section{Introduction}
\label{sec: introduction}
Large language models (LLMs) have garnered significant acclaim and success across various domains and applications \citep{llama3, gpt3, t5}. With ongoing advancements, the scope and complexity of released LLMs have witnessed exponential growth, with some LLMs encompassing $>$50B parameters \citep{deepseekv3, opt, bloom}. This remarkable upscaling introduces considerable challenges, particularly when deploying these models or granting their accessibility to users with constrained resources. To address these challenges, weight-only post-training quantization (PTQ) has emerged as an effective approach.
 
PTQ methods can generally be classified into three categories: statistic-based, gradient-based, and codebook-based approaches. Statistic-based methods \citep{spqr, awq, gptq} determine the quantization grid based on the weight value distribution, whereas gradient-based methods \citep{omniquant, affinequant} optimize the quantization grid with some calibration samples. Codebook-based methods~\citep{aqlm, gptvq, squeezellm, anyprecision} cluster similar weight elements to the shared quantized centroids, employing non-uniform quantization and pushing the limits to extremely low bit levels. However, these methods continue to struggle with low-bit quantization and the presence of outliers in weights, leading to significant performance degradation.

Another challenge PTQs encounter is their limited support for finetuning, which is crucial for adapting LLMs to various downstream tasks. Finetuning LLMs is computationally expensive due to their large scale and the need to cache activations and store optimizer states. PTQ, which compresses LLMs, appears to be a promising approach for finetuning as it reduces the memory requirements for loading LLMs. However, most quantization techniques use a round-to-nearest operation, which does not support gradient backpropagation. Typically, parameter-efficient methods \citep{qlora, loftq, apiq} are employed to train the added parameters while keeping the quantized LLMs frozen, bypassing this limitation. However, this fine-tuning approach presents two major drawbacks: (1) freezing quantized LLMs prevents further reduction of quantization errors during finetuning; (2) The low-rank nature of most parameter-efficient methods restricts their expressiveness \citep{loraleans, road}.

In this paper, we propose a simple while effective paradigm that mainly applies \textbf{Clus}tering to \textbf{Comp}ress LLMs, referred to as \textit{ClusComp}. Additionally, ClusComp can function as a parameter and memory-efficient finetuning method. Our preliminary experiments reveal that LLMs are increasingly difficult to quantize, mainly due to the growing frequency of outliers in their weight matrices (\S\ref{sec: method}). Based on this observation, we propose using clustering instead of quantization to compress LLMs, retaining all values in FP16 format to circumvent issues arising from outlier quantization (\S\ref{subsubsec: clustering}). To further reduce compression errors, we minimize block-wise output discrepancies between the compressed and uncompressed blocks, using a limited set of calibration samples (\S\ref{subsubsec: block-wise}). In addition, by incorporating an inexpensive end-to-end recovery finetuning step, we can even push the compression rate to the 1-bit level. Since all parameters remain in FP16 after compression, ClusComp fully supports standard neural training, thus allowing the seamless finetuning of compressed LLMs on various downstream tasks (\S\ref{subsubsec: recovery finetuning}).

We begin by evaluating the effectiveness of ClusComp in the context of model compression across 2 language modeling tasks and 6 zero-shot reasoning tasks. ClusComp consistently surpasses various baselines at 2-4 levels, even achieving a perplexity of $<$13 at the 2-bit level on WikiText2 \citep{wikitext} for all LLMs (\S\ref{subsec: compression}). Following recovery finetuning, ClusComp's performance at 2-bit and 1-bit levels approaches that of the FP16 model, with an accuracy of 57.8 vs 68.6 for the 2-bit Llama-3-8B and 51.4 vs 75.4 for the 1-bit Llama-3-70B (\S\ref{subsec: limit}). Additionally, ClusComp demonstrates its utility as a parameter-efficient ($<1\%$) and memory-efficient (42GB for Llama-3-70B) finetuning method, outperforming quantization-based and memory-efficient finetuning approaches, while matching the performance of full finetuning (\S\ref{subsec: finetuning}).

\begin{figure*}[t]
  \centering
  \includegraphics[width=0.9\textwidth]{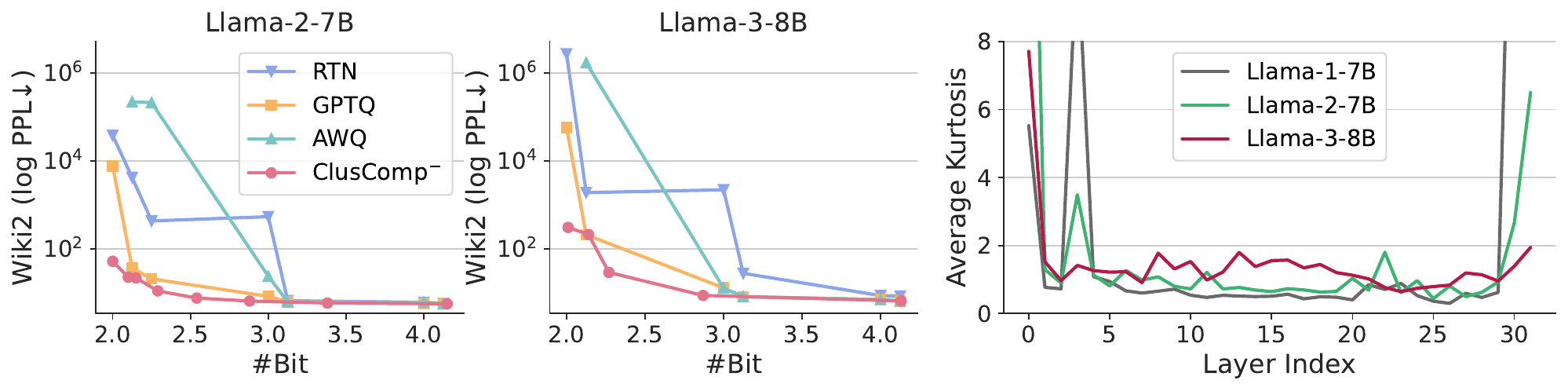}
  \caption[]{The Llama series becomes progressively harder to quantize. \textbf{Left} \& \textbf{Middle:} From Llama-2 to Llama-3, all methods struggle more with lower-bit quantization, though ClusComp$^-$ is the least affected. \textbf{Right:} From Llama-1 to Llama-3, the average weight kurtosis across most layers increases.}
  \label{fig: kurtosis}
\end{figure*}

\section{Related Works}
\label{sec: backgraound}

\textbf{Quantization} refers to the process of converting floating-point values into discrete levels, thereby reducing the bit-width required and minimizing memory consumption during model loading. Taking the symmetric uniform quantization as an example, a weight matrix $\bm{W}$ is quantized as follows:
\begin{align}
    \bm{W}_q = \text{clamp}(\nint{\frac{\bm{W}}{s}}, -2^{b-1}, 2^{b-1}-1) \nonumber
\end{align}
where the scale factor $s=\frac{\text{max}(|\bm{W}_{\text{min}}|, |\bm{W}_{\text{max}}|)}{2^{b}-1}$, $b$ denotes the bit-width, and $\nint{}$ represents the round-to-nearest (RTN) operation. Since the quantization grid is uniform, its effectiveness is contingent on the distribution of the weight values. In cases where the weight matrix contains a significant number of outliers or is quantized to lower bit-widths, the resulting quantization error may be substantial.

\textbf{Post-training quantization} (PTQ) methods, such as GPTQ \citep{gptq}, AWQ \citep{awq}, and OmniQuant \citep{omniquant}, apply quantization to a model after training with minimal computational resources. However, these approaches, which rely on uniform quantization, are significantly impacted by the presence of outliers in the weight matrices. Recent methods \citep{spqr, pbllm, billm} address this challenge by retaining salient weights in FP16 format, thereby maintaining strong performance at lower bit widths. Nonetheless, these mixed-precision quantization techniques require specially optimized CUDA kernels to either enhance or preserve inference speed. Closely related to our proposed method, ClusComp, are works such as GPTVQ \citep{gptvq}, QuIP\# \citep{quipsharp} and SqueezeLLM \citep{squeezellm} which implement quantized codebooks for non-uniform quantization, achieving state-of-the-art performance for ultra-low-bit quantization. ClusComp, however, differs in two significant ways: (1) The codebook in ClusComp is stored in FP16, offering additional advantages for subsequent recovery training and finetuning; (2) While other methods face limitations similar to those in VAE-like approaches \citep{vae}, where large and high-dimensional codebooks are infeasible due to mode collapse, ClusComp circumvents this issue. Our fixed-code design allows us to utilize a codebook size of $2^{16}$ in 4-16D without encountering such difficulty.

\textbf{Finetuning} is crucial for adapting LLMs to various domains and applications. Quantization, which reduces model size, is theoretically more conducive to finetuning. However, directly finetuning a quantized model is not a standard approach, as RTN does not support gradient backpropagation. Finetuning using a straight-through estimator (STE) \citep{ste} is relatively under-explored and may lead to catastrophic forgetting \citep{pvtuning}. Previous works \citep{qa-lora, apiq, qlora, liao2023parameter} propose freezing the quantized model while updating newly added LoRAs \citep{lora}. However, these approaches suffer from two key limitations: (1) Freezing the quantized model prevents the mitigation of quantization errors during finetuning. Moreover, not all quantization methods are suitable for finetuning. Popular techniques like GPTQ and QLoRA, while widely used, exhibit significant quantization errors below 4-bit. (2) The expressiveness of LoRA is constrained by its bottleneck design \citep{loraleans, road}. In contrast, ClusComp, where all parameters are maintained in high precision, inherently supports seamless finetuning. Additionally, updating the codebook in ClusComp results in modifying all parameters in the weight matrices, even offering superior performance compared to full finetuning while maintaining a similar number of trainable parameters as LoRA.

\section{ClusComp}
\label{sec: method}

\label{subsec: pilot}
\textbf{Pilot study.} Before introducing ClusComp, we present a key observation from our experiments on different Llama series. As in Figure \ref{fig: kurtosis} (Left \& Middle), when reducing the bit-width, the performance of various quantization methods (RTN, GPTQ, and AWQ) follows a similar trend: Llama-3 \citep{llama3} proves more challenging to quantize than Llama-2 \citep{llama2}.

We hypothesize that the increasing quantization difficulty stems from a higher frequency of outliers in Llama-3's linear layers. Since these methods rely on uniform quantization, they are particularly sensitive to outliers in weight matrices. To test this, we analyzed the kurtosis of weight matrices—a well-established metric for detecting outliers \citep{outliers}. As shown in Figure \ref{fig: kurtosis} (Right), we observe: (1) Higher kurtosis at the beginning and end of all models; (2) A consistent increase in kurtosis from Llama-1 to Llama-3 across most layers, indicating more frequent outliers. This trend likely explains the growing quantization challenge and suggests that future Llama models may be even harder to quantize.\footnote{Further discussion and proof are in \S\ref{sec: quantization difficulty}.}

\textit{Since quantization performance is affected by outliers, could an alternative compression approach involve storing all weight values in FP16 rather than applying quantization?}

The first idea that comes to mind is clustering, where similar weight values are represented by a single shared value. This approach enables model compression while maintaining all weight values in FP16. In the following, we introduce three ClusComp variants that leverage clustering for LLM compression: \textit{ClusComp$^-$}, which applies clustering alone; \textit{ClusComp}, which enhances ClusComp$^-$ with block-wise error minimization; and \textit{ClusComp$^+$}, which further refines the compressed model through next-token prediction.

\subsection{ClusComp\texorpdfstring{$^-$}:: Clustering}
\label{subsubsec: clustering}
Considering a weight matrix $\bm{W} \in \mathbb{R}^{d_{\text{in}} \times d_{\text{out}}}$, direct clustering along either dimension of $\bm{W}$ is suboptimal as it leads to a significant reconstruction error, particularly due to the large values of $d_{\text{in}}$ and $d_{\text{out}}$ in LLMs. To mitigate this issue, we reshape $\bm{W}$ into a set of lower-dimensional vectors, denoted as $\bm{W}' = \{\bm{w}_1, \bm{w}_2, \dots, \bm{w}_k\}$, where each $\bm{w}_i \in \mathbb{R}^{g}$ and $k=\frac{d_{\text{in}} \cdot d_{\text{out}}}{g}$.\footnote{In cases where the dimensions are not divisible, zero-padding is applied to $\bm{W}$. In PyTorch, the matrix is reshaped as $\bm{W}' = \bm{W}$.transpose(1, 0).view(-1, g), where transposing $\bm{W}$ offers slightly better performance.} The goal is to partition $\bm{W}'$ into $n$ clusters $\{C_1, C_2, \dots, C_n\}$ by solving the following optimization problem:
\begin{align}
    \text{argmin}_{\{C_1, C_2, \dots, C_n\}} \sum_{j=1}^n \sum_{\bm{w}_i \in C_j} ||\bm{w}_i - \bm{c}_j||^2 \nonumber
\end{align}
where $\bm{c}_j \in \mathbb{R}^g$ denotes the centroid of cluster $C_j$. This clustering problem is well-established in the machine learning literature and can be iteratively addressed using K-means \citep{kmeans} with the Expectation-Maximization (EM) algorithm:
\begin{itemize}
    \item \textbf{E-step}: Each vector $\bm{w}_i$ is assigned to the cluster whose centroid $\bm{c}_j$ minimizes the Euclidean distance, i.e., $C_j^{(t)} = \{\bm{w}_i: ||\bm{w}_i - \bm{c}_j^{(t)}||^2 \leq ||\bm{w}_i - \bm{c}_l^{(t)}||^2 \quad \forall l\}$.
    \item \textbf{M-step}: The centroid of each cluster is updated as the mean of the vectors assigned to that cluster, i.e., $\bm{c}_j^{(t+1)} = (\sum_{\bm{w}_i \in C_j^{(t)}} \bm{w}_i)/|C_j^{(t)}|$.
\end{itemize}

Upon completion, two key elements are obtained for each weight matrix: (1) a codebook $\bm{C} = \{\bm{c}_1, \bm{c}_2, \dots, \bm{c}_n\} \in \mathbb{R}^{g \times n}$ that contains all centroids, and (2) a set of codes $\bm{q} = \{q_1, q_2, \dots, q_k\} \in \{1, 2, \dots, n\}^k$ that records the assignment of each vector $\bm{w}_i$ to the closest centroid, where $q_i = q(\bm{w}_i) = j$ if $\bm{c}_j = \text{argmin}_{\bm{c}_l \in \bm{C}} ||\bm{w}_i - \bm{c}_l||^2$. Using the codes $\bm{q}$ and the codebook $\bm{C}$, the weight matrix $\bm{W}'$ can be reconstructed as $\hat{\bm{W}'} = \{\bm{c}_{q_1}, \bm{c}_{q_2}, \dots, \bm{c}_{q_k}\}$. In PyTorch~\citep{pytorch}, the linear layer is adapted as in Listing \ref{lst: linear}.

\textit{Remark}: ClusComp, when applied solely with clustering, is referred to as ClusComp$^-$. In this configuration, only the weight matrices (no calibration samples) are utilized, leading to substantial memory efficiency, with a mere 2GB memory consumption for both 7B and 70B LLM on 1 GPU, as shown in Figure \ref{fig: overall}. Moreover, this process can be considerably accelerated with more GPUs, as the clustering of different matrices is independent.

\begin{table}[t]
    \begin{center}
    \scriptsize
    \begin{tabular}{ccccc}
    \toprule
    \textbf{Setting} & \textbf{\#Params. for Codes} & \textbf{\#Params. for Codebook} & \textbf{$\bar{b}$} \\
    \midrule
    g4n65500 & 4.19M & 0.26M (1.55\%) & 4.25 \\
    g6n65500 & 2.80M & 0.39M (2.32\%) & 3.04 \\
    g9n65500 & 1.86M & 0.59M (3.52\%) & 2.34 \\
    \bottomrule
    \end{tabular}
    \caption{Bits for $\bm{W}$ with $d_{\text{in}}, d_{\text{out}} = 4096$ (16.78M).}
    \label{tab: bit example}
    \end{center}
\end{table}

\subsection{Estimate model size}
After clustering, it is sufficient to store the codes $\bm{q} \in \{1, 2, \dots, n\}^k$ and codebook $\bm{C} \in \mathbb{R}^{g \times n}$. Unlike prior works \citep{gptvq, aqlm, quipsharp}, we don't quantize the codebook; instead, we store it in the FP16 format. The bit-width required for the codes depends on the range of $n$. To maintain efficiency, we set $n < 2^{16}$ and use unsigned 16-bit integers to represent the codes. Thus, the average bits-per-parameter is calculated as:
\begin{align}
    \bar{b} &= \frac{\text{size in bits}}{\text{number of parameters}} \nonumber \\
    &= \frac{16 \cdot k + 16 \cdot g \cdot n}{d_{\text{in}} \cdot d_{\text{out}}} = \frac{16}{g} + \frac{16 \cdot g \cdot n}{d_{\text{in}} \cdot d_{\text{out}}} \label{eq: bit}
\end{align}

In the right-most of Equation (\ref{eq: bit}), the first term corresponds to the bit-width allocated to the codes, and the second term corresponds to the bit-width of the codebook. E.g., for a linear layer $\bm{W}$ with $d_{\text{in}} = d_{\text{out}} = 4096$, clustering with $g = 4$ and $n = 2^{16} - 1$ results in $\bar{b} \approx 4 + 0.25 = 4.25$. This demonstrates that the majority of the bit-width is allocated to the codes, which is a primary reason for constraining $n < 2^{16}$, so we can use 16 instead of 32-bit integers to represent the code. Further reducing $n$ to a smaller range leads to fewer centroids, which in turn increases reconstruction error. More settings can be found in Table \ref{tab: bit example} and \ref{tab: wiki ppl}.

\textit{Remark}: While we express the model size in terms of bits-per-parameter, it is important to note that no quantization is applied in ClusComp. Instead, we reduce the number of parameters in $\bm{W}$ from $d_{\text{in}} \cdot d_{\text{out}}$ to $k + g \cdot n$. Utilizing bits-per-parameter allows for a direct comparison between ClusComp and quantization-based methods. Surprisingly, ClusComp$^-$, which only incorporates the clustering step without employing any calibration data, already surpasses RTN, GPTQ and AWQ, as illustrated in Figure \ref{fig: kurtosis} (Left \& Middle). 

\subsection{ClusComp: Block error minimization}
\label{subsubsec: block-wise}
Block-wise error minimization (block-wise reconstruction or knowledge distillation) has emerged as a standard, efficient and effective approach to reducing quantization error \citep{aqlm, quipsharp, gptvq, omniquant, apiq}. To further mitigate the compression error caused by clustering, we incorporate block-wise error minimization into ClusComp$^-$ using a limited set of calibration samples, expressed as: 
\begin{align} 
    \text{argmin}_{\bm{C} s}||\mathcal{F}(\bm{W} s, \bm{X}) - \mathcal{F}(\bm{C} s, \bm{q} s, \bm{X}')|| \label{eq: block-wise} 
\end{align} 
Here, $\mathcal{F}$ denotes a Transformer block \citep{transformer}, $\bm{W}$s represent the weight matrices in the uncompressed block, and $\bm{C}$s and $\bm{q}$s denote the codebooks and codes in the compressed block. $\bm{X}$ refers to the input of the uncompressed block, which is also the output from the previous uncompressed block, while $\bm{X}'$ is the input to the compressed block, originating from the output of the preceding compressed block. For the first block, we have $\bm{X} = \bm{X}'$. Block-wise error minimization is memory-efficient as it only requires loading two blocks into the GPU simultaneously.

\textit{Remark}: In Equation (\ref{eq: block-wise}), we only train the codebook $\bm{C}$s while keeping the codes $\bm{q}$s fixed as indices. This design offers two key advantages: (1) It enhances data efficiency. As illustrated in Table \ref{tab: bit example}, the majority of parameters are represented by the codes. Training both the codebooks and codes with a limited number of calibration samples (128) leads to overfitting; (2) More importantly, training the codes with a large number of centroids ($2^{16}$) can result in mode collapse \citep{llamagen, vae}. Since the codes already exhibit a uniform distribution after clustering (see Figure \ref{fig: index dist}), keeping the codes fixed indicates that all centroids in the codebook can be trained uniformly. Such a code-fixed design is also applied to the following recovery and finetuning step. Combining both clustering and block-wise error minimization steps, we term this method ClusComp.

\subsection{ClusComp\texorpdfstring{$^+$}:: Recovery and finetuning}
\label{subsubsec: recovery finetuning}
We present the adapted linear layer for ClusComp in Listing \ref{lst: linear}, which can be seamlessly integrated as a replacement for the original linear layer in LLMs. As the codebook is represented in FP16, this new layer inherently supports training without requiring additional tricks, like STE.

\textbf{Recovery training.} The compressed LLMs can be further trained by predicting the next token to recover information lost due to compression. This is achieved by finetuning the codebook parameters. This form of training is memory-efficient in two distinct ways: (1) Since the LLM is already compressed, loading it onto the GPU consumes less memory compared to the FP16 version; (2) As illustrated in Table \ref{tab: bit example}, the parameters in the codebook account for $<5\%$ of the total parameters in the FP16 version, making the training both parameter-efficient and memory-efficient. We refer to ClusComp with recovery training as ClusComp$^+$.

\textbf{Finetuning.} Like recovery training, finetuning the compressed LLM on downstream tasks can also be performed efficiently. Unlike QLoRA \citep{qlora}, which freezes the quantized LLM and trains only the LoRA \citep{lora} modules, finetuning the codebook alone eliminates the need for this additional constraint. This approach offers two key advantages over QLoRA: (1) Freezing the quantized LLM prevents mitigation of quantization errors, whereas finetuning the codebook can further address compression errors for downstream tasks; (2) The low-rank bottleneck of LoRA limits its expressiveness \citep{loraleans}. In contrast, finetuning the codebook is analogous to adapting the entire high-rank weight matrix, providing greater flexibility and expressiveness.

\section{Experiments}

\begin{figure*}[t]
  \centering
  \includegraphics[width=0.85\textwidth]{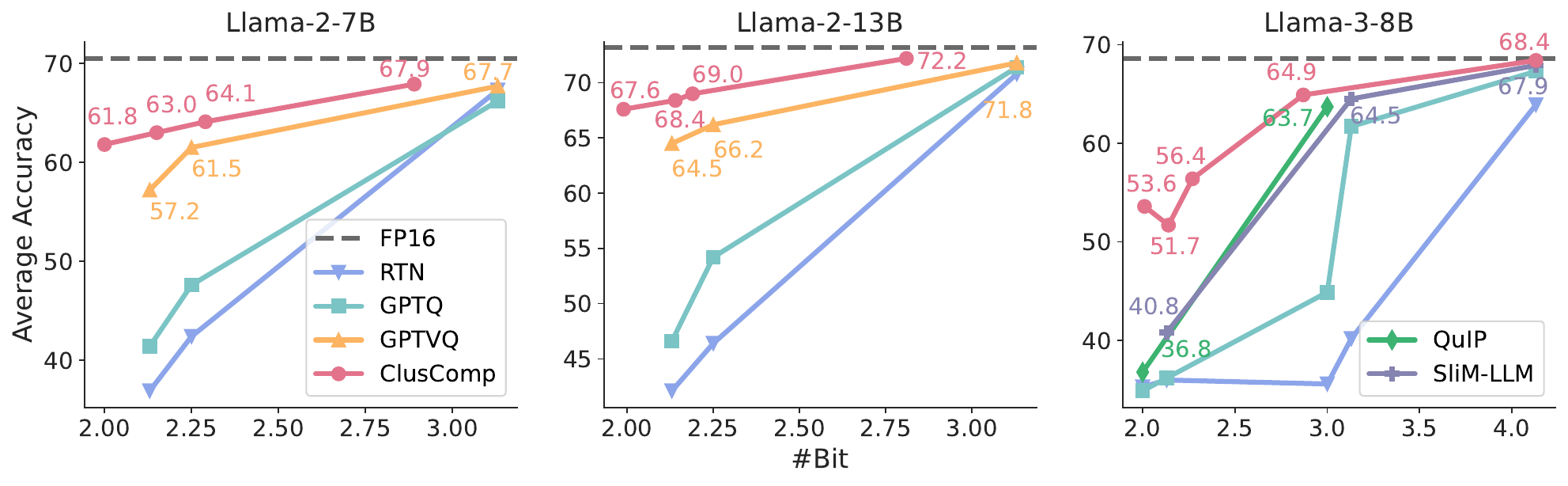}
  \caption[]{Average zero-shot accuracy over 5 or 6 commonsense reasoning tasks, only including competitive baselines. Please refer to Table \ref{tab: commonsense 1} and \ref{tab: commonsense 2} for detailed numbers and the full comparison.}
  \label{fig: commonsense}
\end{figure*}

\begin{table*}[t]
    \begin{center}
    \tiny
    \begin{tabular}{lrrcccccc}
    \toprule
    \textbf{Method} & \multicolumn{1}{c}{\textbf{\#Bit}} & \multicolumn{1}{c}{\textbf{AI2D} $\uparrow$} & \textbf{ChartQA} $\uparrow$ & \textbf{DocVQA} $\uparrow$ & \textbf{MMBench} $\uparrow$ & \textbf{Avg} $\uparrow$ & \textbf{MME} (cog / per) $\uparrow$ \\
    \cmidrule(lr){3-7}
    \textbf{LLaVA-Next-8B}  & 16.00 & 71.7 & 69.2 & 78.2 & 72.2 & 72.8 & 1965.1 (376.8 / 1588.3) \\
    \midrule
    GPTQ & 4.13 & \textbf{70.7} & 67.4 & 77.4 & 71.0 & 71.6 & 1895.0 (331.6 / 1563.4) \\
    AWQ & 4.13 & 70.6 & 68.0 & 77.2 & \textbf{71.1} & 71.7 & 1888.4 (325.7 / 1562.7) \\
    \rowcolor{green!8}
    \textbf{ClusComp} & 4.13 & 70.0 & \textbf{68.7} & \textbf{77.6} & \textbf{71.1} & \textbf{71.8} & \textbf{1915.7} (322.1 / 1593.6) \\
    \cmidrule(lr){1-8}
    GPTQ & 3.13 & 66.2 & 65.1 & \textbf{75.6} & 67.4 & 68.6 & 1831.8 (290.1 / 1541.7) \\
    AWQ & 3.13 & 67.7 & 65.4 & 74.4 & \textbf{68.0} & 68.9 & 1840.3 (298.6 / 1541.7) \\
    \rowcolor{green!8}
    \textbf{ClusComp} & \textbf{2.87} & \textbf{68.7} & \textbf{65.8} & 74.8 & 67.7 & \textbf{69.3} & \textbf{1872.6} (331.1 / 1541.5) \\
    \cmidrule(lr){1-8}
    GPTQ & 2.13 & 0.0 & 0.0 & 0.0 & 0.0 & 0.0 & 0.0 (0.0 / 0.0) \\
    AWQ & 2.13 & 0.0 & 0.0 & 0.0 & 0.0 & 0.0 & 0.0 (0.0 / 0.0) \\
    \rowcolor{green!8}
    \textbf{ClusComp} & 2.14 & \textbf{53.9} & \textbf{53.1} & \textbf{56.7} & \textbf{50.1} & \textbf{53.5} & \textbf{1673.0} (294.6 / 1378.4) \\
    \bottomrule
    \end{tabular}
    \caption{Zero-shot multimodal evaluation, with baseline results from \citet{llama3quant}.}
    \label{tab: multimodal}
    \end{center}
\end{table*}

\subsection{Compression results}
\label{subsec: compression}
\textbf{LLMs and evaluation.} We evaluate ClusComp on widely adopted LLM families: Llama-1-7B, Llama-2-7B/13B/70B and Llama-3-8B/70B \citep{llama, llama2, llama3}. We measure the performance of compressed LLMs on zero-shot and language modeling tasks. For zero-shot evaluation, we apply 6 tasks from lm-eval v0.4.4 \citep{lmeval}, i.e. PIQA \citep{piqa}, ARC-e/c \citep{arc}, BoolQ \citep{boolq}, HellaSwag \citep{hellaswag} and WinoGrande \citep{winogrande}. For language modeling, we report the perplexity on the whole test set of WikiText2 \citep{wikitext} and on 256 samples from the validation set of C4 \citep{c4} with a sequence length of 2048 as our baselines. We also apply ClusComp to LLaVA-Next-8B \citep{llava-next}, and evaluate it on 5 multimodal tasks from lmms-eval v0.2.3 \citep{lmms_eval} to show its broad applicability, i.e. AI2D \citep{ai2d}, ChartQA \citep{chartqa}, DocVQA \citep{docvqa}, MMBench \citep{mmbench} and MME \citep{mme}.

\textbf{Baselines.} Here we primarily compare ClusComp with three categories of baselines: (1) statistic-based methods without neural training, including vanilla RTN, GPTQ \citep{gptq}, AWQ \citep{awq}, and PB-LLM \citep{pbllm};\footnote{ClusComp$^{-}$ is also a statistic-based method.} (2) gradient-based methods with neural training (such as block-wise distillation), including OmniQuant \citep{omniquant}, AffineQuant \citep{affinequant}, and SliM-LLM$^+$ \citep{slim-llm}; and (3) quantized codebook-based methods, including QuIP \citep{quip} and GPTVQ \citep{gptvq}. All baseline results are directly borrowed from the original works or their follow-up works.

\textbf{Settings.} We begin by applying K-means clustering to the weight matrices in all linear layers, referring to this method as ClusComp$^-$. Next, we use 128 samples from the WikiText2 training set to minimize block-wise error through codebook training only, which we denote as ClusComp. It is important to note that the majority of the aforementioned baselines utilize resources comparable (GPU memory and the number of calibration samples) to those used in ClusComp. Please refer to \S \ref{sec: experimental details} for all experimental details in this section.

\textbf{Results.} The main language modeling results are shown in Figure \ref{fig: overall}. At the 4-bit level, all methods are comparable. At bit-widths $<4$, ClusComp consistently outperforms all baselines, with a larger gap for a lower bit-width. Notably, even at the 2-bit level, ClusComp's perplexity remains within a functional range, $<13$ on WikiText2. Figure \ref{fig: commonsense} presents the zero-shot evaluation results, where ClusComp again consistently surpasses all baselines across different bit-widths. Furthermore, ClusComp exhibits significantly less sensitivity to bit-width variations, as indicated by the flatter slope of its accuracy curve.

We also compress the Llama-3-8B backbone in LLaVA-Next-8B, and report the zero-short performance in Table \ref{tab: multimodal}. On average, ClusComp continues to outperform both GPTQ and AWQ, while using a comparable or even lower number of bits. A particularly noteworthy observation occurs at the 2-bit level, where none of the baselines produce correct outputs, whereas ClusComp retains strong performance. In comparison to the 2-bit results for Llama-3-8B in Figure \ref{fig: commonsense} (Right), this suggests that quantizing multimodal models presents unique challenges, warranting further study.

\begin{table}[t]
    \begin{center}
    \tiny
    \begin{tabular}{lrc|lrc}
    \toprule
    \textbf{Method} & \textbf{\#Bit} & \textbf{Avg. Acc}$\uparrow$ & \textbf{Method} & \textbf{\#Bit} & \textbf{Avg. Acc}$\uparrow$ \\
    \midrule
    \textbf{Llama-2-7B} & 16.00 & 64.8 & \textbf{Llama-2-13B} & 16.00 & 67.8 \\
    QuiP\# & 2.02 & 57.5 & QuiP\# & 2.01 & 61.5 \\
    AQLM & 2.02 & 56.5 & AQLM & 1.97 & 60.6 \\
    \rowcolor{green!8}
    \textcolor{gray!80}{ClusComp} & \textcolor{gray!80}{2.00} & \textcolor{gray!80}{56.6} &  \textcolor{gray!80}{ClusComp} & \textcolor{gray!80}{1.99} & \textcolor{gray!80}{62.0} \\
    \rowcolor{green!8}
    \textbf{ClusComp$^+$} & 2.00 & \textbf{57.8} & \textbf{ClusComp$^+$} & 1.99 & \textbf{63.1} \\
    \midrule
     \textbf{Llama-3-8B} & 16.00 & 68.6 & \textbf{Llama-3-70B} & 16.00 & 75.4 \\
    QuiP & 2.00 & 36.8 & QuIP & 2.00 & 48.7 \\
    PB-LLM & 2.00 & 38.8 & PB-LLM & 1.70 & 44.1 \\
    DB-LLM & 2.00 & 51.8 & BiLLM & 1.10 & 44.2 \\
    \rowcolor{green!8}
    \textcolor{gray!80}{ClusComp} & \textcolor{gray!80}{2.01} & \textcolor{gray!80}{53.6} &  \textcolor{gray!80}{ClusComp} & \textcolor{gray!80}{1.14} & \textcolor{gray!80}{39.7} \\
    \rowcolor{green!8}
    \textbf{ClusComp$^+$} & 2.01 & \textbf{57.8} & \textbf{ClusComp$^+$} & 1.14 & \textbf{51.4} \\
    \bottomrule
    \end{tabular}
    \caption{Zero-shot accuracy on ultra-low-bit LLMs. Please refer to Table \ref{tab: limit detailed number} for accuracy on various tasks.}
    \label{tab: limit}
    \end{center}
\end{table}

\subsection{Push the limit of model compression}
\label{subsec: limit}
We further enhance the performance of 2-bit LLMs and extend the compression boundary to the 1-bit level through efficient recovery training. This is achieved by optimizing the codebook parameters in an end-to-end next-token prediction task.

\textbf{Baselines.} We include BiLLM \citep{billm}, which performs effectively at the 1-bit compression level. Additionally, three more resource-intensive PTQ baselines are considered: AQLM \citep{aqlm}, which utilizes a larger number of calibration samples (4-16M tokens); QuIP\# \citep{quipsharp} and DB-LLM \citep{dbllm}, both of which employ model-wise distillation and a larger number of calibration samples (24-48M tokens).

\textbf{Settings.} ClusComp employs only 0.3M tokens for its compression. In this experiment, we further finetune the compressed LLM generated by ClusComp through end-to-end training, optimizing the codebook parameters using 16M tokens from a subset of the RedPajama dataset \citep{redpajama}. This extended method is referred to as ClusComp$^+$.

\textbf{Results.} We report the zero-shot accuracy of ultra-low-bit LLMs in Table \ref{tab: limit}. On both Llama-2-7B and 13B, ClusComp already performs comparably to, or surpasses AQLM and QuiP\#. With minimal recovery training, ClusComp$^+$ consistently outperforms these baselines on average, with the performance gap increasing for larger LLMs, indicating the robust scalability of ClusComp$^+$. On Llama-3-8B, ClusComp already exceeds all baselines, and ClusComp$^+$ further widens this margin. On Llama-3-70B, ClusComp$^+$ achieves remarkable accuracy at the 1-bit level. Furthermore, when comparing the improvements from ClusComp to ClusComp$^+$ across different Llama series, a notably larger performance gain is observed on the Llama-3 models, underscoring the effectiveness of ClusComp$^+$ on LLMs with more outliers.

\subsection{Finetuning quality}
\label{subsec: finetuning}
We can finetune the compressed LLMs on downstream tasks by only training the codebooks.

\begin{table}[t]
    \begin{center}
    \scriptsize
    \begin{tabular}{lrrcc}
    \toprule
     \multirow{2}{*}{\textbf{Method}} & \multirow{2}{*}{\textbf{Bit}} & \multirow{2}{*}{\textbf{\#Trained}} & \textbf{MMLU} & \textbf{AGIEval}  \\
    & & & 5-shot $\uparrow$ & 3-shot $\uparrow$ \\
    \midrule
    \textbf{Llama-2-7B} & 16.00 & - & 45.9 & 25.7 \\
    Full FT & 16.00 & 100\% & 45.7 & 27.0  \\
    \cmidrule(lr){1-5}
    LoRA ($r=128$) & 16.00 & 4.9\% & 45.5 & 24.7 \\
    GaLore & 16.00 & 100.0\% & 45.5 & 24.4 \\
    LISA & 16.00 & 100.0\% & 46.2 & 26.1 \\
    \rowcolor{green!8}
    \textbf{ClusComp} & 4.15 & 0.9\% & \textbf{47.0} & \textbf{26.5} \\
    \rowcolor{green!8}
    \textbf{ClusComp} & 2.88 & 1.4\% & 45.1 & 25.6 \\
    \rowcolor{green!8}
    \textbf{ClusComp} & 2.00 & 1.4\% & 30.7 & 21.8 \\
    \bottomrule
    \end{tabular}
    \caption{ClusComp performance against efficient finetuning, with baseline results from~\citet{lisa}.}
    \label{tab: agieval}
    \end{center}
\end{table}

\textbf{General-domain finetuning.} We finetune the compressed LLMs on a new version of Alpaca, i.e. Alpaca-GPT4 \citep{alpaca4}, and evaluate them on both MMLU and AGIEval \citep{agieval}. Here we mainly compare ClusComp to some memory-efficient finetuning methods that fully finetune the FP16 version, i.e. GaLore \citep{galore} and LISA \citep{lisa}. As shown in Table \ref{tab: agieval}, finetuning the compressed LLMs at the 4-bit level from ClusComp outperforms all memory-efficient finetuning methods, and rivals full finetuning. In addition, the finetuned LLMs can be used in a low bit, friendly for inference.

The superior finetuning performance can be attributed to three key advantages of ClusComp: (1) ClusComp introduces smaller compression errors; (2) Unlike QLoRA, where compressed LLMs are frozen during finetuning, ClusComp allows for the model to remain unfrozen by training the codebook parameters, enabling further mitigation of compression errors; (3) The low-rank design of LoRA limits its expressiveness. In contrast, updating the codebook in ClusComp is analogous to updating a high-rank weight matrix. In addition, the fixed-code design allows uniform training of all centroids, providing greater expressiveness and can even rival full finetuning.

\begin{figure*}[t]
  \centering
  \includegraphics[width=0.8\linewidth]{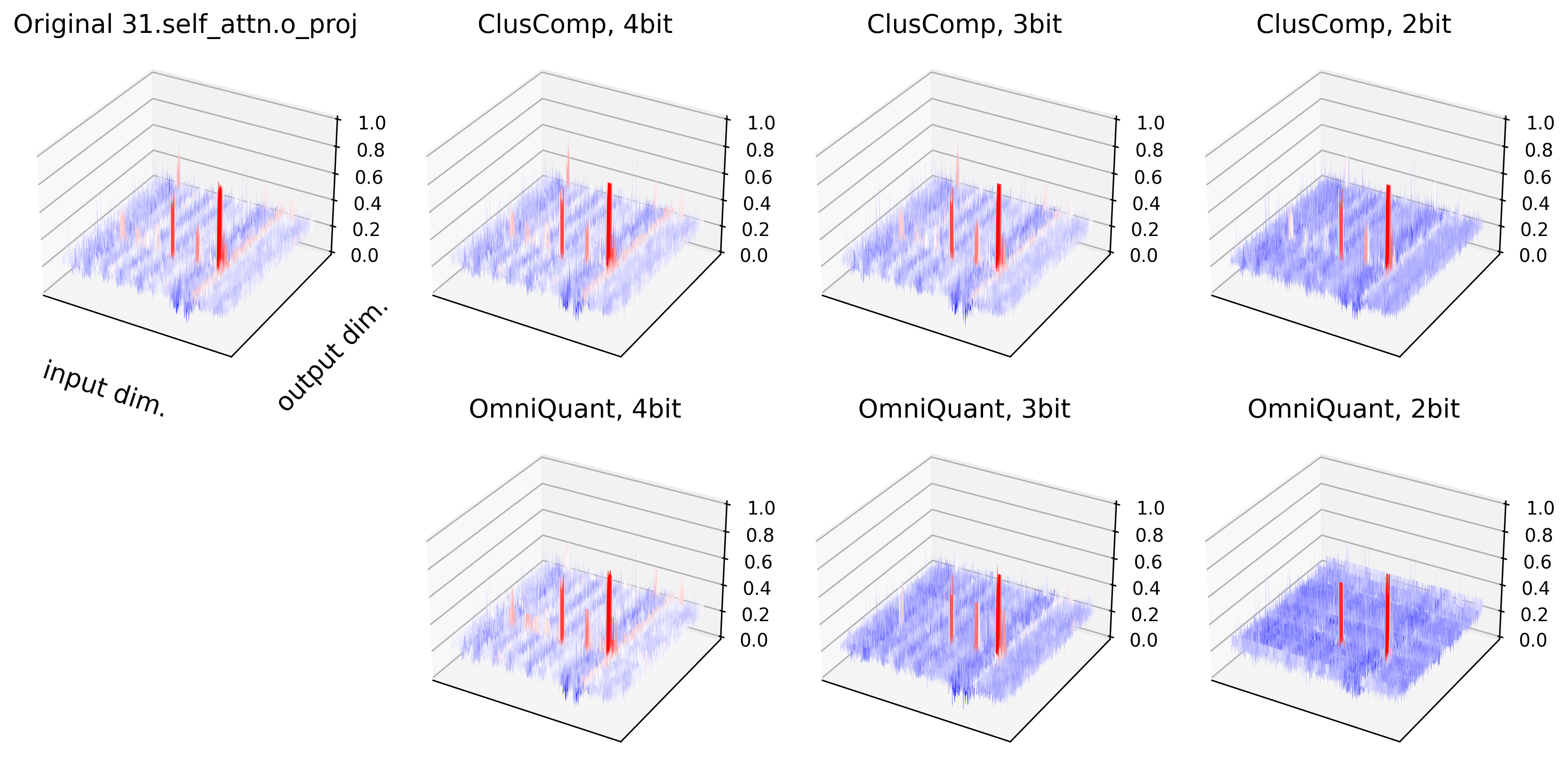}
  \caption[]{Weight patterns of a cherry-picked layer in Llama-2-7B. Darker red and blue indicate larger and smaller weight values, respectively. Please refer to \S\ref{subsec: weight visualization} for the visualization setting and the visualization of more layers.}
  \label{fig: weight vis}
\end{figure*}

\begin{figure}[t]
  \centering
  \includegraphics[width=0.85\linewidth]{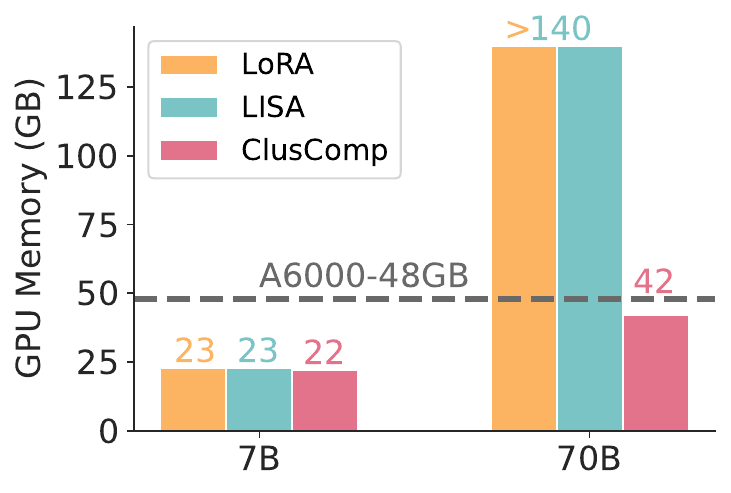}
  \caption[]{Memory consumption for recovery training or finetuning. 4-bit LLMs are used for ClusComp here.}
  \label{fig: mem}
\end{figure}

\section{Discussion}
\label{sec: discussion}

\subsection{Memory efficiency} 
Figure \ref{fig: mem} illustrates the memory efficiency of ClusComp during training, with a batch size of 1 and a sequence length of 1024. For the 70B LLM, we apply gradient checkpointing (this is not used for the 7B LLM), while omitting any additional memory-saving techniques.

Finetuning the LLM compressed by ClusComp demonstrates memory efficiency in two key ways: (1) The compressed LLM requires less memory for loading onto the GPU compared to the FP16 model; and (2) Only the codebook parameters, which contain a limited number of trainable parameters ($<1\%$), are updated, as detailed in Table \ref{tab: agieval}. Consequently, the optimizer state size remains small. ClusComp can serve not only as a model compression technique but also as an effective method for both memory-efficient and parameter-efficient finetuning.

\subsection{Weight distribution} 
In Figure \ref{fig: weight vis}, we compare ClusComp's weight distribution to OmniQuant's. Notably, OmniQaunt applies uniform quantization. Overall, ClusComp’s weight distribution in different bit levels can better simulate the original weight distribution, explaining ClusComp's effectiveness.

\begin{table}[t]
    \begin{center}
    \scriptsize
    \begin{tabular}{rrcc}
    \toprule
    \textbf{\#Bits for Codebook} & \textbf{Avg. \#Bit} & \textbf{Llama-2-7B} & \textbf{Llama-2-70B} \\
    \midrule
    - & 16.00 & 1.00$\times$ & 1.00$\times$ \\
    16.00 & 2.01 & 1.29$\times$ & 1.17$\times$ \\
    8.00 & 1.92 & 1.31$\times$ & 1.19$\times$ \\
    4.00 & 1.86 & 2.10$\times$ & 1.81$\times$ \\
    \bottomrule
    \end{tabular}
    \end{center}
    \caption{Throughput comparison between ClusComp and the FP16 version. Originally, the codebook in ClusComp remained in FP16. For further inference speedup, we can uniformly quantize it to a lower bit level.}
    \label{tab: throughput}
\end{table}

\subsection{Inference latency}
Originally, the data type of the codebook $\bm{C}$ is in 16-bit, which facilitates our following recovery training and finetuning step. However, if the codebook can be further quantized, we have two additional advantages: (1) The model size can be slightly reduced (Only slightly since the majority of bits is allocated to the code $\bm{q}$.); (2) The inference can speed up, since the codebook becomes smaller. In sum, we can keep the codebook in 16-bit if we want to do the recovery training or finetuning. If we are only interested in inference, we can further quantize the codebook.

As shown in Table \ref{tab: codebook quantization}, the performance doesn't change if we quantize the codebook from 16-bit to 8-bit. When quantizing the codebook to 4-bit, the perplexity slightly increases, but is still comparable to the best baseline at the 4-bit level and outperforms the best baseline at the 2-bit level. However, if we further quantize the codebook to the 2-bit level, the perplexity increases significantly. Therefore, we can safely quantize the codebook to 8-bit or 4-bit. 

In Table \ref{tab: throughput}, we show the throughput of ClusComp. When the codebook stays in the FP16 format, the throughput is only slightly better than the FP16 LLM, because non-uniform compression is not friendly to hardware. To enhance the inference, we can further quantize the codebook to a lower bit level. When the codebook is quantized to a 4-bit level, we can roughly achieve a 2$\times$ speedup.

\begin{table*}[t]
    \centering
    \small
    \begin{subtable}[t]{0.3\textwidth}
        \centering
        \begin{tabular}{crr}
        \toprule
        \textbf{Setting} & \textbf{\#Bit} & \textbf{Wiki2} \\
        \midrule
        g4 & 4.14 & \textbf{5.67} \\
        g5 & 3.38 & 5.90 \\
        g6 & 2.89 & 6.54 \\
        g7 & 2.54 & 7.64 \\
        g8 & 2.29 & 11.10 \\
        \bottomrule
        \end{tabular}
        \caption[]{$n=65500$. Perplexity changes smoothly.}
    \end{subtable}
    \hfill
    \begin{subtable}[t]{0.3\textwidth}
        \centering
        \begin{tabular}{lrr}
        \toprule
        \textbf{Setting} & \textbf{\#Bit} & \textbf{Wiki2} \\
        \midrule
        g7n16384 & 2.35 & \textbf{23.14} \\
        g7n4096 & 2.30 & 9.4e2 \\
        \cmidrule(lr){1-3}
        g8n65500 & 2.29 & \textbf{11.10} \\
        g8n50000 & 2.15 & 21.90 \\
        g8n4096 & 2.02 & 5.4e3 \\
        \bottomrule
        \end{tabular}
        \caption[]{Same $g$. Perplexity changes dramatically.}
    \end{subtable}
    \hfill
    \begin{subtable}[t]{0.3\textwidth}
        \centering
        \begin{tabular}{lrr}
        \toprule
        \textbf{Setting} & \textbf{\#Bit} & \textbf{Wiki2} \\
        \midrule
        g7n16384 & 2.35 & 23.14 \\
        g7n4096 & 2.30 & 9.4e2 \\
        g8n65500 & 2.29 & \textbf{11.10} \\ 
        \bottomrule
        \end{tabular}
        \caption[]{Same $g$. Perplexity changes dramatically.}
    \end{subtable}
    \caption[]{Ablation study of ClusComp$^-$ on the number of clusters $n$ and the cluster dimension $g$ in the codebook reveals that $n$ plays a more significant role in the performance of the quantized LLM. \textbf{(a)} The perplexity remains relatively stable with variations in $g$, although changes in $g$ lead to substantial differences in the bit requirement, as most bits are used to store the codes $\bm{q}$. Specifically, smaller $g$ values result in larger $\bm{q}$. Refer to Table \ref{tab: bit example} for detailed examples. \textbf{(b)} In contrast, perplexity is highly sensitive to changes in $n$. Adjusting $n$ causes only a minor change in the bit requirement, as storing the codebook is memory-efficient. \textbf{(c)} For comparable bit budgets, $n$ has a greater impact on performance than $g$.\label{tab: ablation}}
\end{table*}

\subsection{Ablation study on the group size and number of clusters}
\label{subsec: abaltion on g and n}
We conducted experiments to determine whether the number of clusters $n$ or the cluster dimension $g$ has a greater impact on the performance of quantized LLMs. As shown in Table \ref{tab: ablation}, increasing $n$ positively affects performance more than reducing $g$. This finding underpins our choice of $n \approx 2^{16}$. However, while $n$ plays a crucial role in enhancing performance, selecting $n > 2^{16}$ would necessitate using 32-bit storage for the code $\bm{q}$, substantially increasing the bits-per-parameter and adversely affecting memory efficiency. Therefore, we always choose $n < 2^{16}$.

\subsection{More results} 
Due to page limit, we recommend readers to (1) Appendix \ref{subsec: more finetuning results} for more finetuning results with a comparison to QLoRA, LoftQ, QA-LoRA, GPTQ-LoRA and PEQA; (2) Appendix \ref{subsec: more baselines} for comparing ClusComp to SqueezeLLM and AdaDim.

\section{Conclusion}
The newly introduced model compression technique, ClusComp, operates by (1) independently applying clustering to the weight matrices to produce both the codebook and corresponding codes, (2) reducing compression error through block-wise knowledge distillation, and (3) enhancing model performance via efficient recovery finetuning. Comprehensive experiments demonstrate its effectiveness as a compression method at 1-4 bit levels, while also showcasing its parameter and memory efficiency for finetuning, with a competitive performance with full finetuning.

\section*{Limitations}
We only evaluate ClusComp on a limited number of tasks with a total number of 7 models due to time and resource limitations. We couldn't guarantee its effectiveness on the other tasks and LLMs, and are still working on including more tasks and models to show its generalization. In addition, applying ClusComp to image generation tasks is not explored in this paper. We leave this exploration to interested readers.

ClusComp only compresses the weight matrices, while leaving the activations in FP16, which might cause OOM error for the long-context generation. However, we can combine ClusComp with activation quantization methods to address this issue.

\section*{Acknowledgments}
We thank eBay Inc. for the computation support. This research was funded in part by the Netherlands Organization for Scientific Research (NWO) under project number VI.C.192.080.

\bibliography{custom}

\clearpage
\appendix
\counterwithin{figure}{section}
\counterwithin{table}{section}

\section{More Results and Discussions}
\label{sec:more results and discussions}

\begin{figure*}[t]
  \centering
  \includegraphics[width=0.8\textwidth]{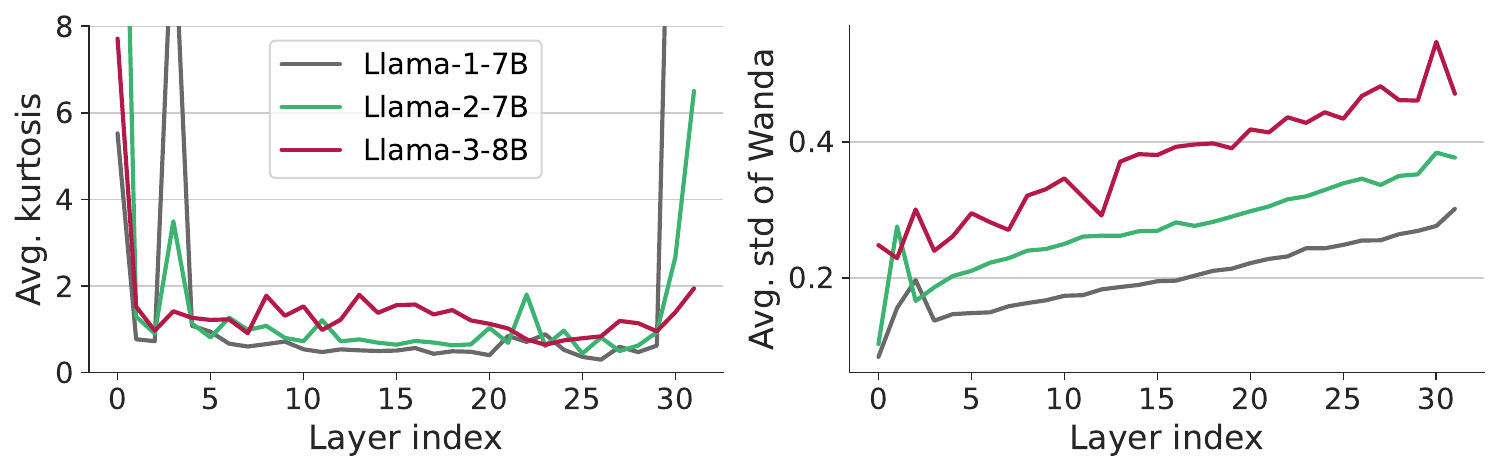}
  \caption[]{From Llama-1 to Llama-3, LLMs exhibit increasing challenges for quantization. \textbf{Left:} The average kurtosis of weights across various layers in the Llama series, previously shown in Figure \ref{fig: kurtosis} (Right). Please refer to Figure \ref{fig: layer kurtosis} for the kurtosis of different layer types. \textbf{Right:} The average standard deviation of the Wanda score across various layers. Please refer to Figure \ref{fig: wanda} for the Wanda scores of different layer types. Both metrics indicate that Llama-3 has higher variance in most layers, reflecting the presence of more outliers and thus greater difficulty for quantization.}
  \label{fig: kurtosis and wanda}
\end{figure*}

\subsection{Quantization difficulty in Llama series}
\label{sec: quantization difficulty}
Previous works \citep{awq, wanda, adadim} suggest that weight patterns can be identified based on activations rather than relying solely on weight magnitudes. We incorporate an additional analysis based on the Wanda score \citep{wanda} distribution to illustrate the increasing challenges of quantization in the Llama series.

Given the weight matrix $\bm{W} \in \mathbb{R}^{d_{out} \times d_{in}}$ of a linear layer and the input activations $\bm{X} \in \mathbb{R}^{NL \times d_{in}}$, where $N$ and $L$ represent the batch and sequence dimensions, the Wanda score $\bm{S} \in \mathbb{R}^{d_{out} \times d_{in}}$ is computed as $\bm{S}_{ij} = |\bm{W}_{ij}|\cdot||\bm{X}_{j}||_2$. A smaller Wanda score within a row of $\bm{W}$ (on a per-output basis) indicates a less significant weight element.

In Figure \ref{fig: kurtosis and wanda} (Right), we present the standard deviation of the Wanda scores across different layers. The results show that Llama-2 exhibits a larger standard deviation compared to Llama-1, while Llama-3 exceeds Llama-2 in this metric. A higher standard deviation reflects a more dispersed Wanda score distribution, indicating that a greater proportion of weight elements are effective and diverse. Consequently, quantization becomes more challenging, as the expanded distribution stretches the quantization grid.

\begin{figure}[t]
  \centering
  \includegraphics[width=0.98\linewidth]{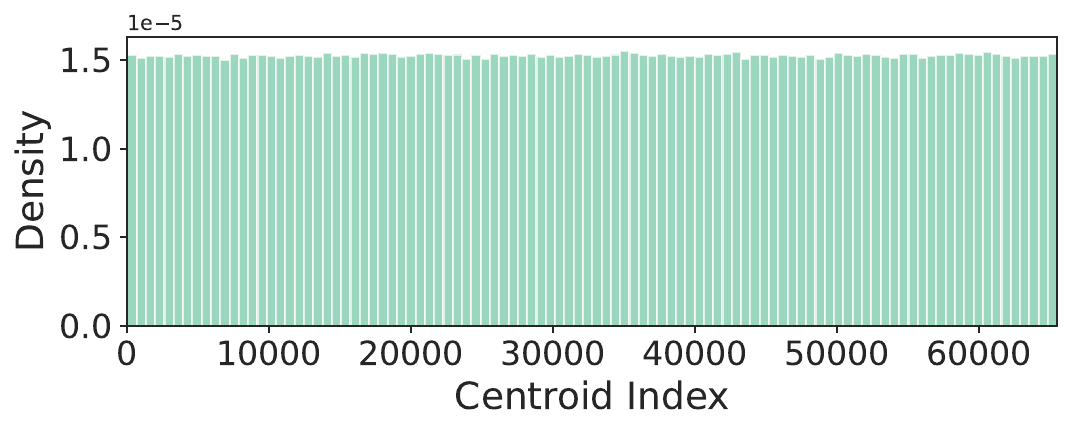}
  \caption[]{Histogram of the codes.}
  \label{fig: index dist}
\end{figure}

\subsection{Code distribution}
As shown in Figure \ref{fig: index dist}, the distribution of codes after clustering is uniform. Training the codebook alone can have a similar effect as full finetuning since all weight values are updated uniformly.

\subsection{Full results on language modeling}
In Table \ref{tab: wiki ppl} and Table \ref{tab: c4 ppl}, we show the full comparison across different LLMs and bit-levels on two language modeling tasks, WikiText2 and C4.

\subsection{Weight visualization}
\label{subsec: weight visualization}
\begin{figure*}[t]
 \centering
 \begin{subfigure}[t]{0.48\linewidth}
     \centering
     \includegraphics[width=\linewidth]{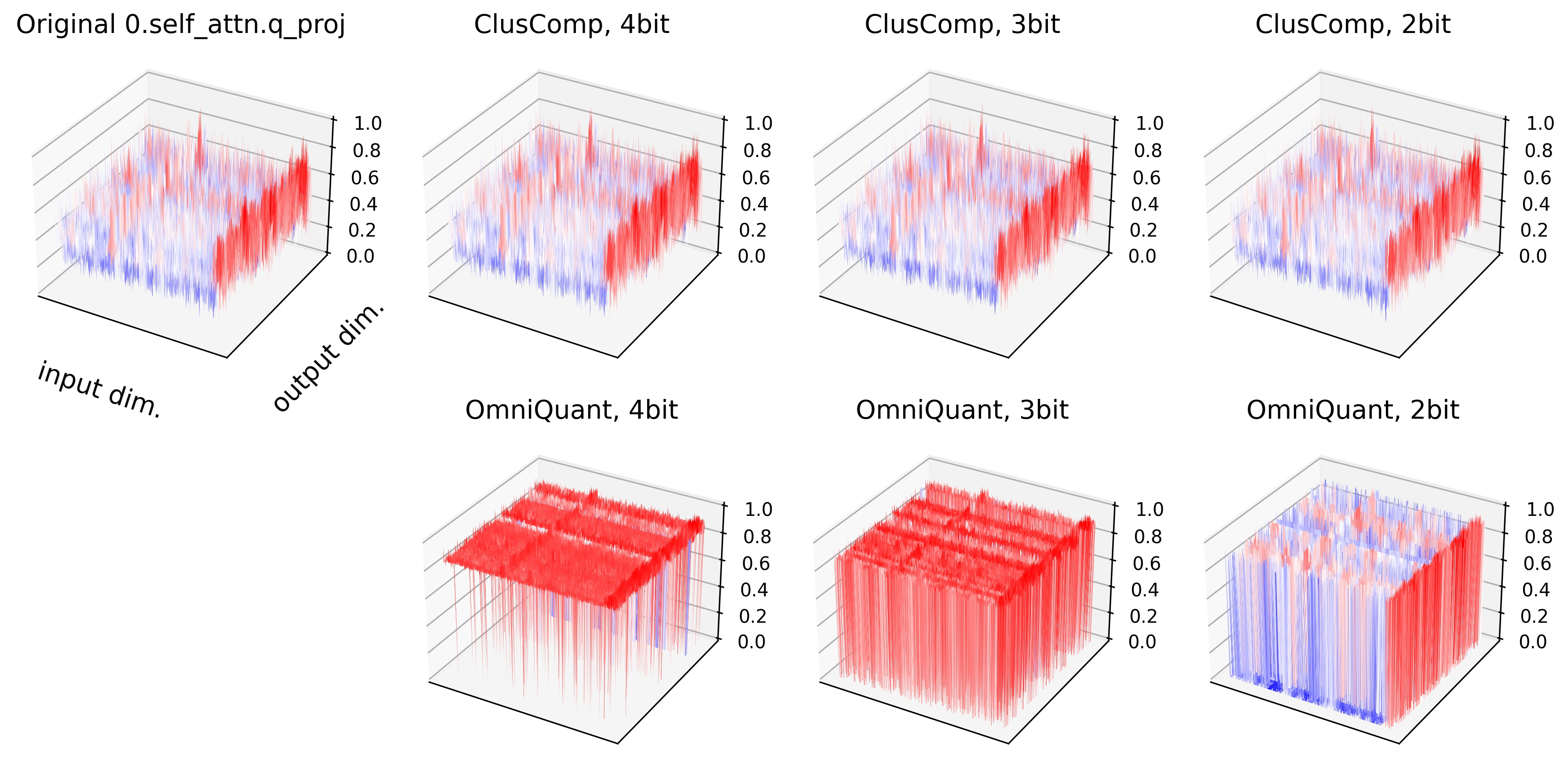}
     \caption{Query projection layer.}
 \end{subfigure}
 \hfill
 \begin{subfigure}[t]{0.48\linewidth}
     \centering
     \includegraphics[width=\textwidth]{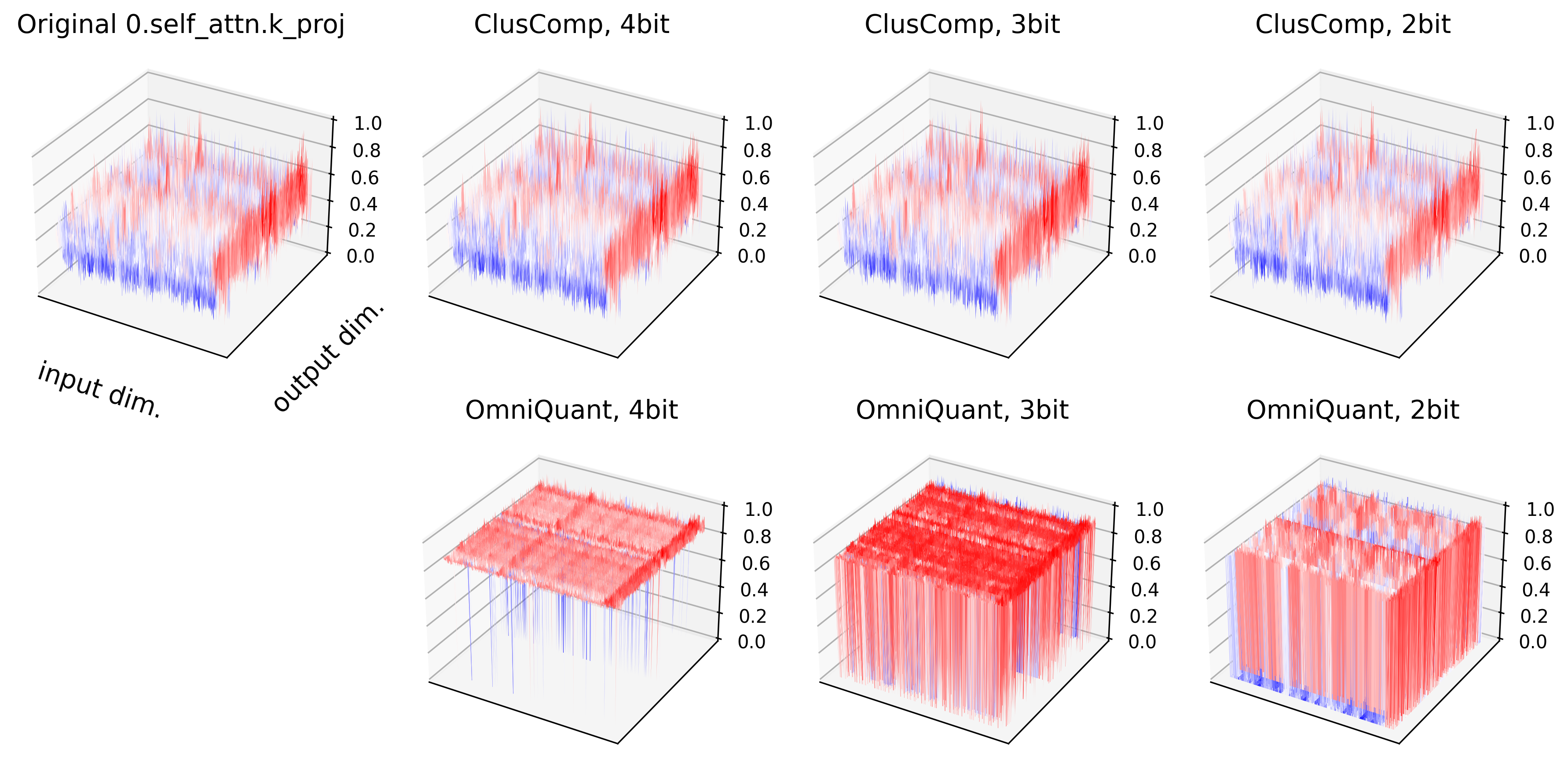}
     \caption{Key projection layer.}
 \end{subfigure}
 \begin{subfigure}[t]{0.48\linewidth}
     \centering
     \includegraphics[width=\textwidth]{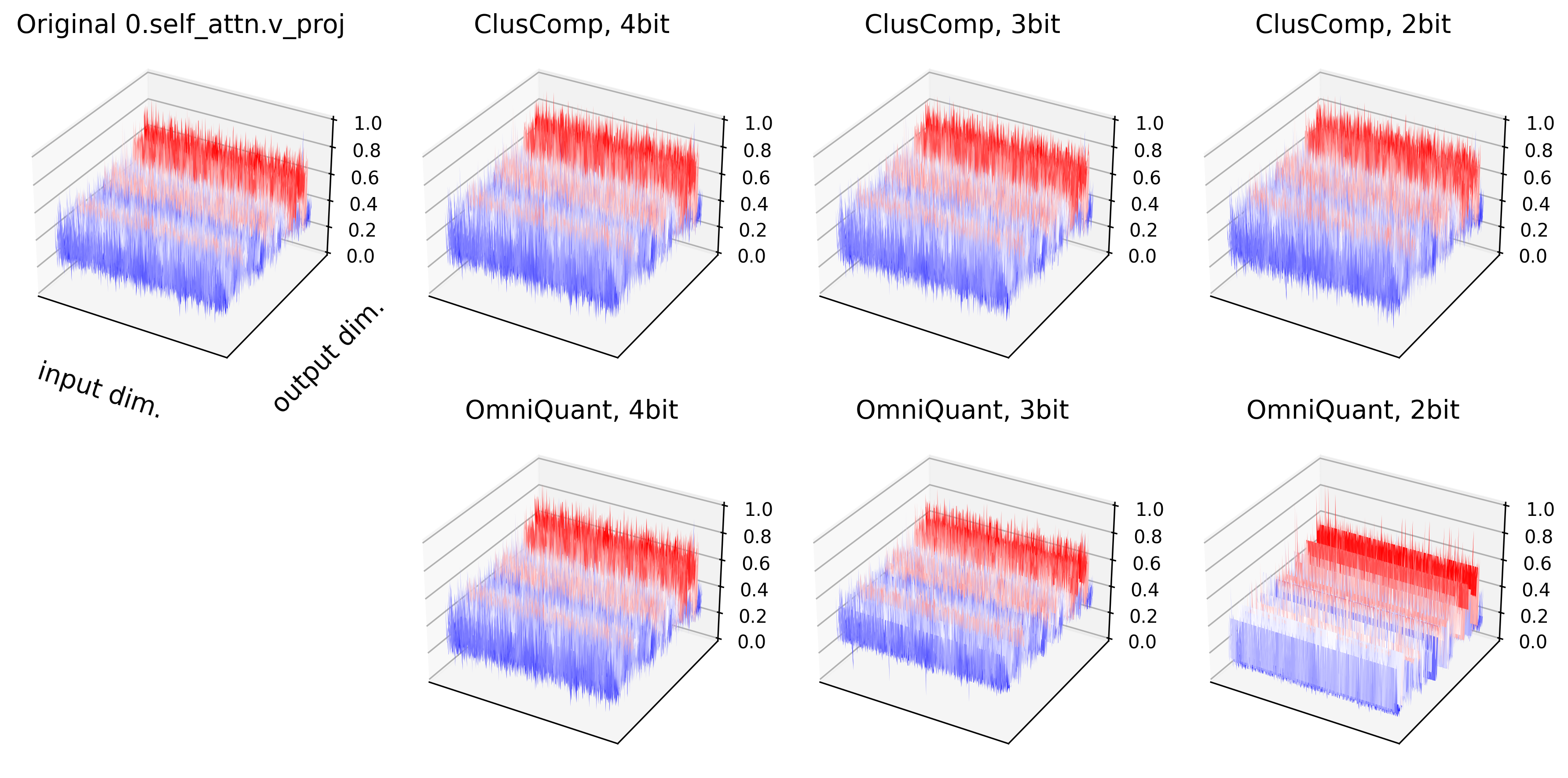}
     \caption{Value projection layer.}
 \end{subfigure}
 \begin{subfigure}[t]{0.48\linewidth}
     \centering
     \includegraphics[width=\textwidth]{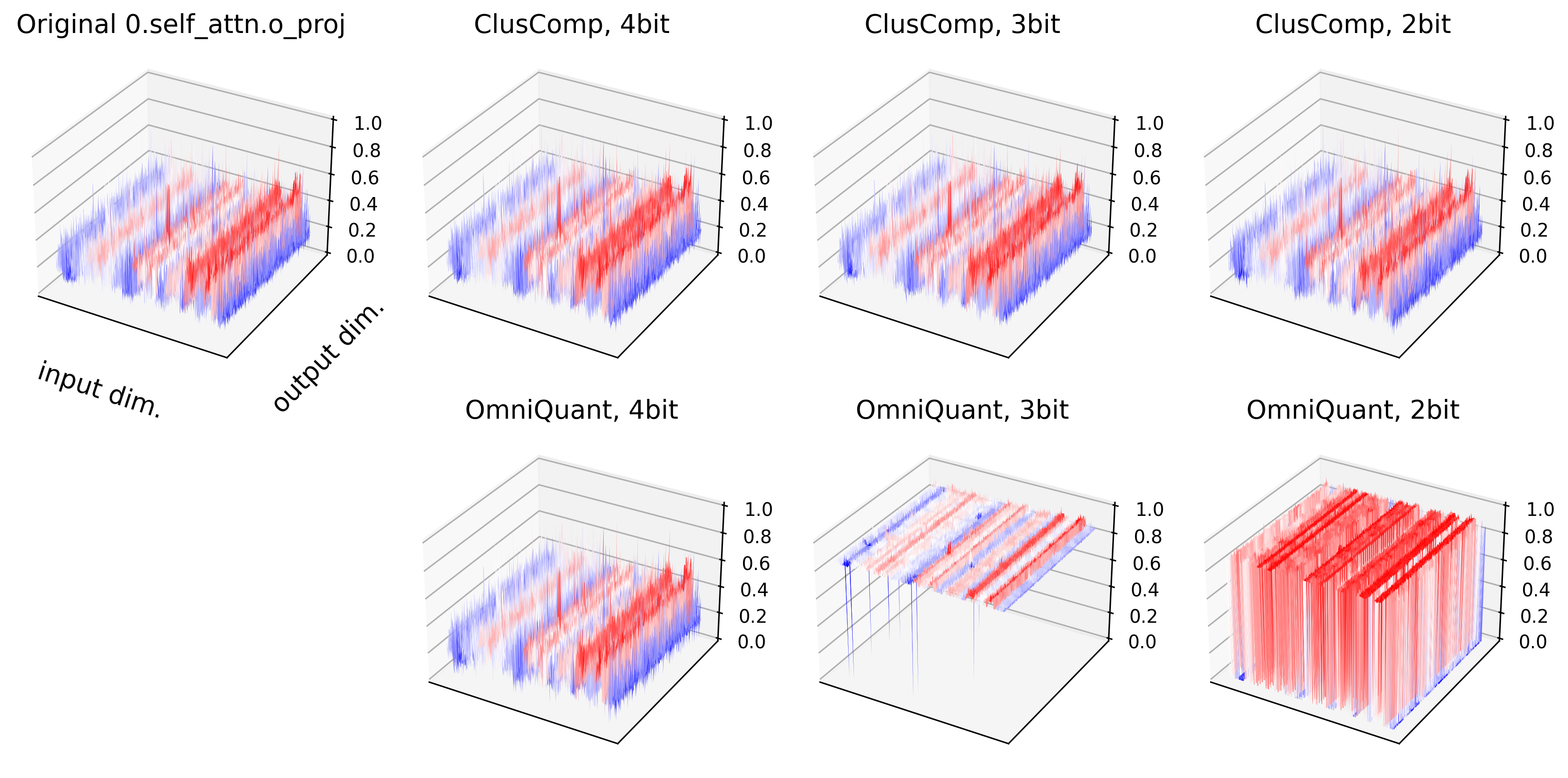}
     \caption{Output projection layer.}
 \end{subfigure}
 \begin{subfigure}[t]{0.48\linewidth}
     \centering
     \includegraphics[width=\textwidth]{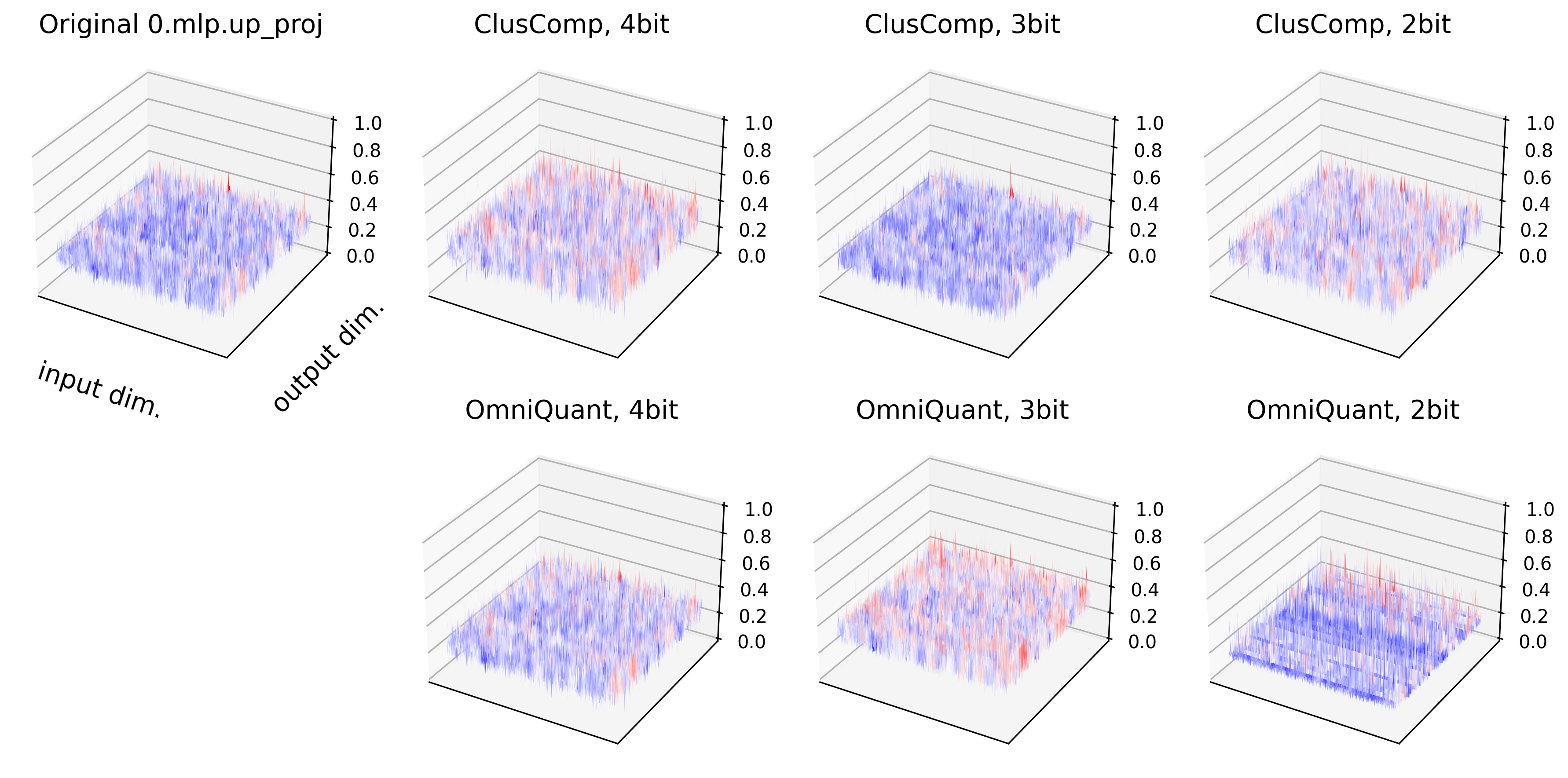}
     \caption{Up projection layer.}
 \end{subfigure}
 \begin{subfigure}[t]{0.48\linewidth}
     \centering
     \includegraphics[width=\textwidth]{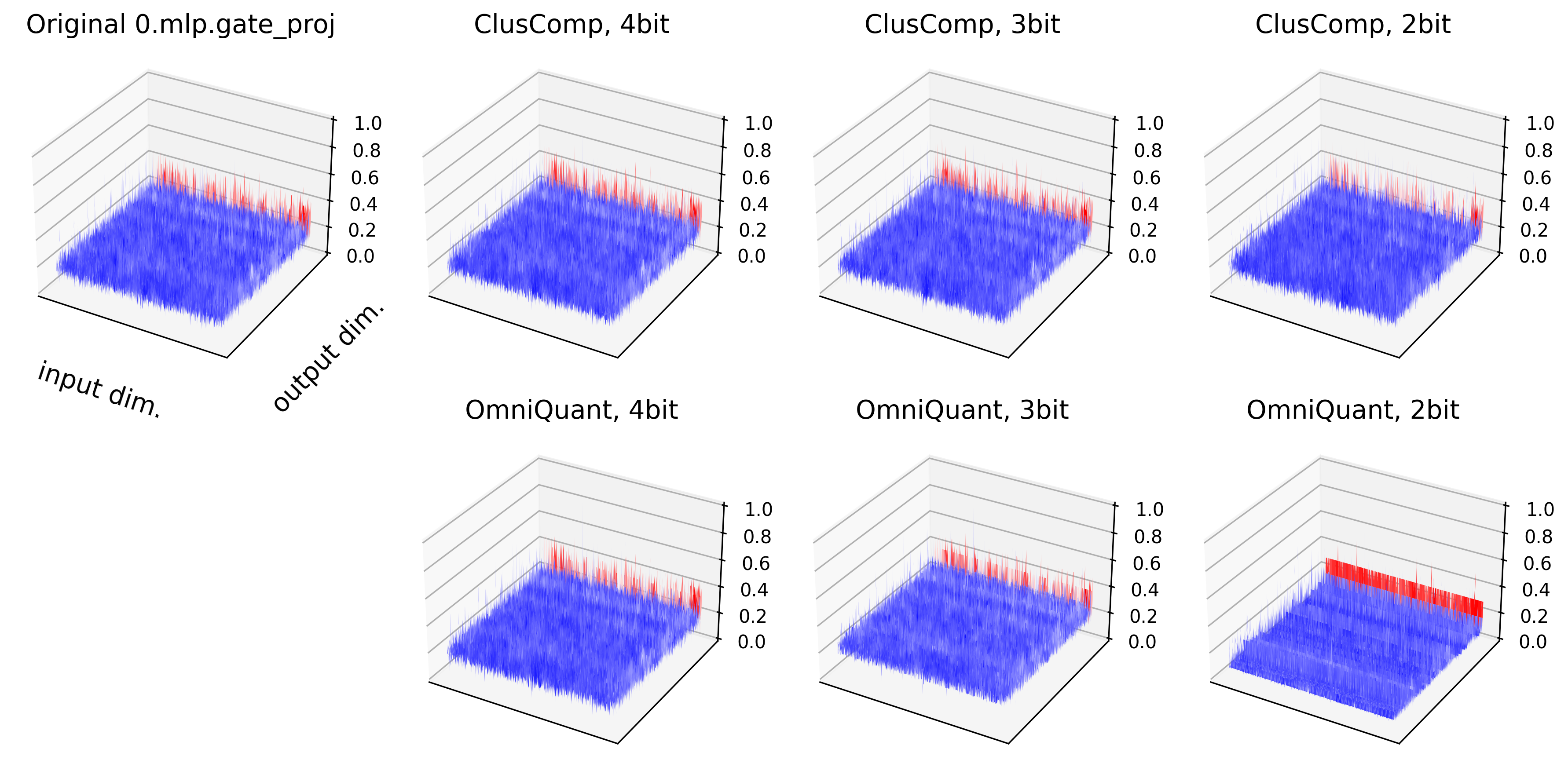}
     \caption{Gate projection layer.}
 \end{subfigure}
 \begin{subfigure}[t]{0.48\linewidth}
     \includegraphics[width=\textwidth]{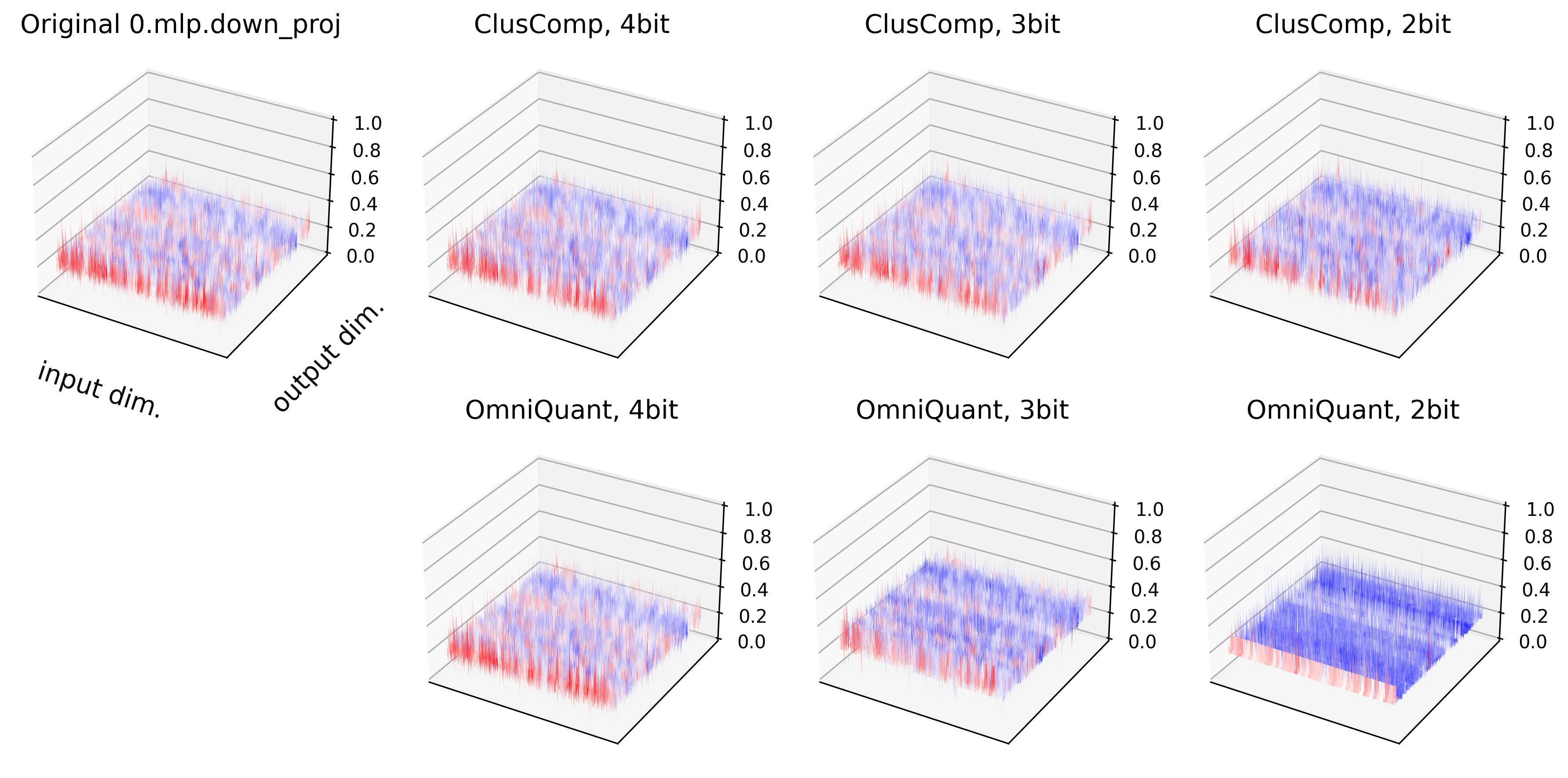}
     \caption{Down projection layer.}
 \end{subfigure}
    \caption[]{Weight patterns of all layers in the first block (0-th block) of Llama-2-7B. Darker red and blue indicate larger and smaller weight values, respectively. ClusComp's weight distribution of different bit levels can better simulate the original weight distribution.}
    \label{fig: first block}
\end{figure*}

\begin{figure*}[t]
 \centering
 \begin{subfigure}[t]{0.48\textwidth}
     \centering
     \includegraphics[width=\textwidth]{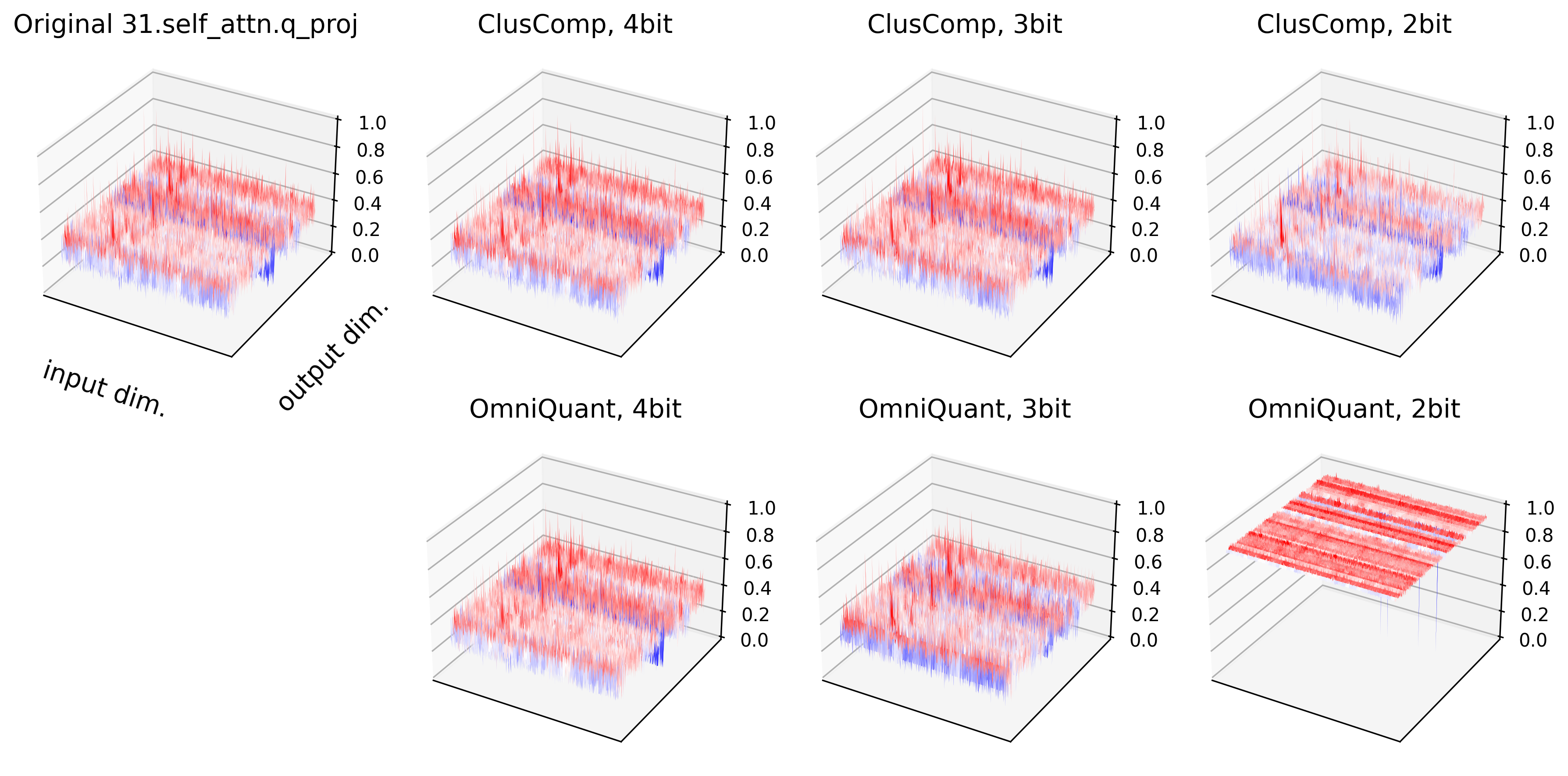}
     \caption{Query projection layer.}
 \end{subfigure}
 \hfill
 \begin{subfigure}[t]{0.48\textwidth}
     \centering
     \includegraphics[width=\textwidth]{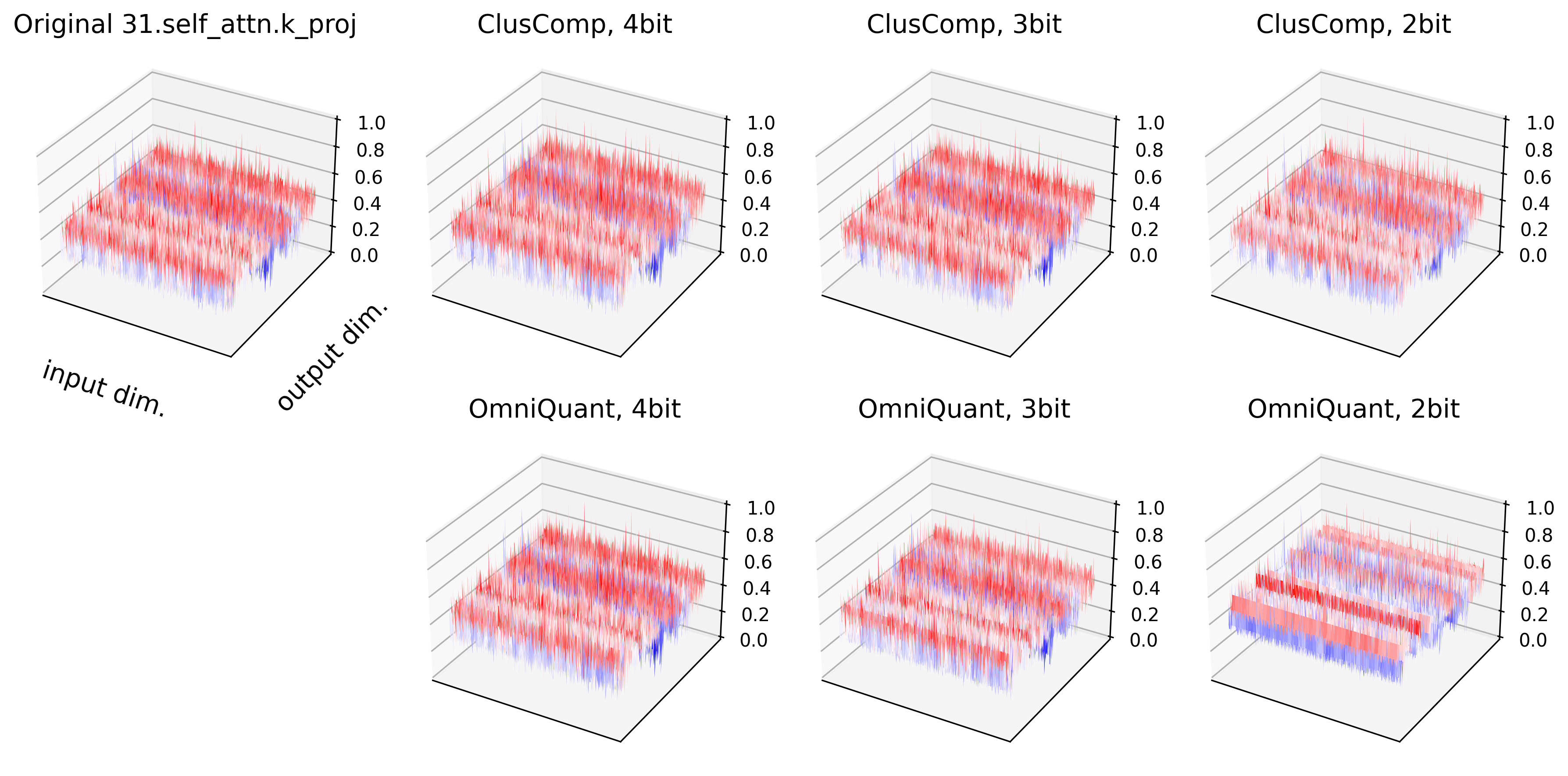}
     \caption{Key projection layer.}
 \end{subfigure}
 \begin{subfigure}[t]{0.48\textwidth}
     \centering
     \includegraphics[width=\textwidth]{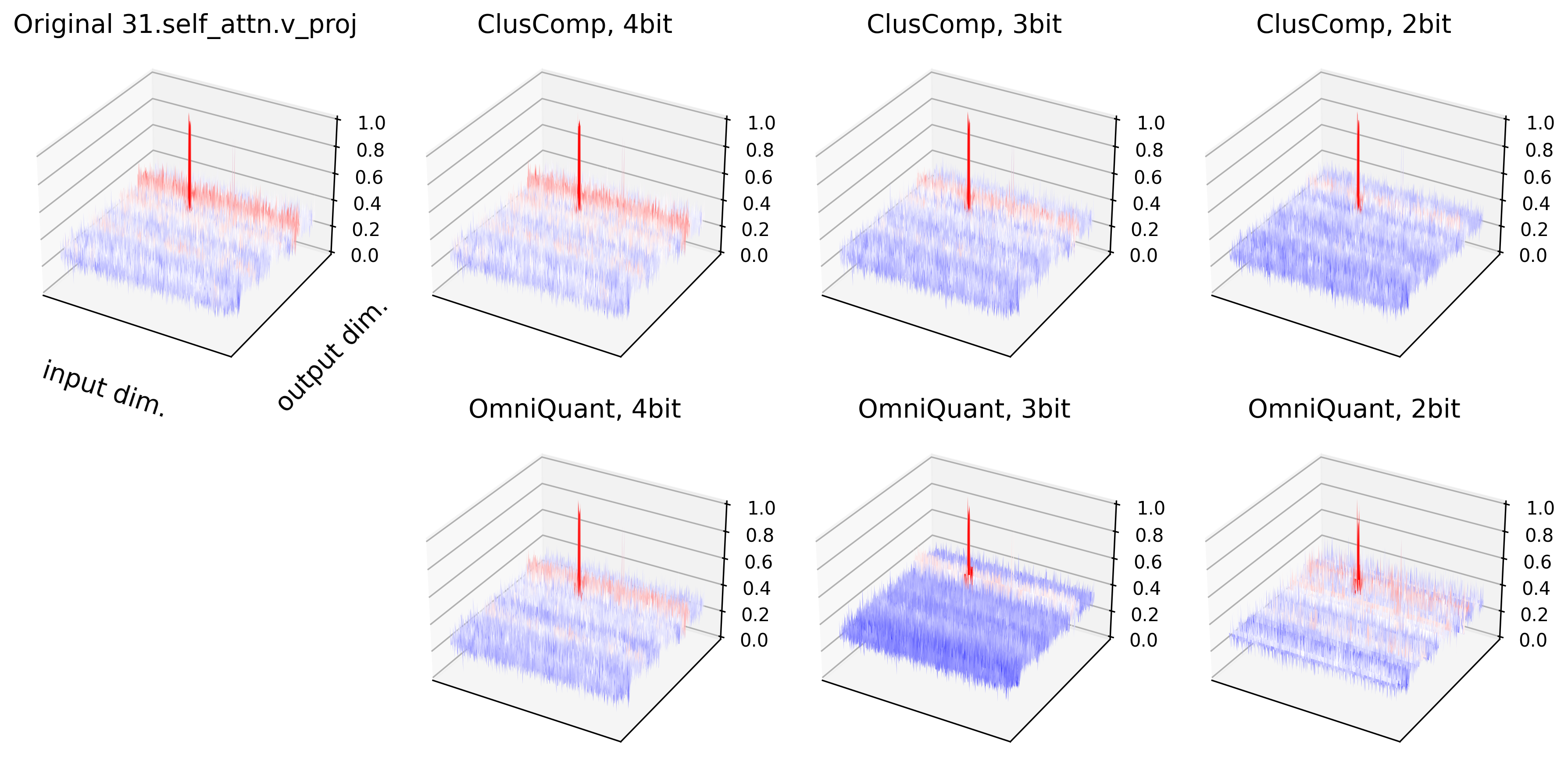}
     \caption{Value projection layer.}
 \end{subfigure}
 \begin{subfigure}[t]{0.48\textwidth}
     \centering
     \includegraphics[width=\textwidth]{figures/o_31.png}
     \caption{Output projection layer.}
 \end{subfigure}
 \begin{subfigure}[t]{0.48\textwidth}
     \centering
     \includegraphics[width=\textwidth]{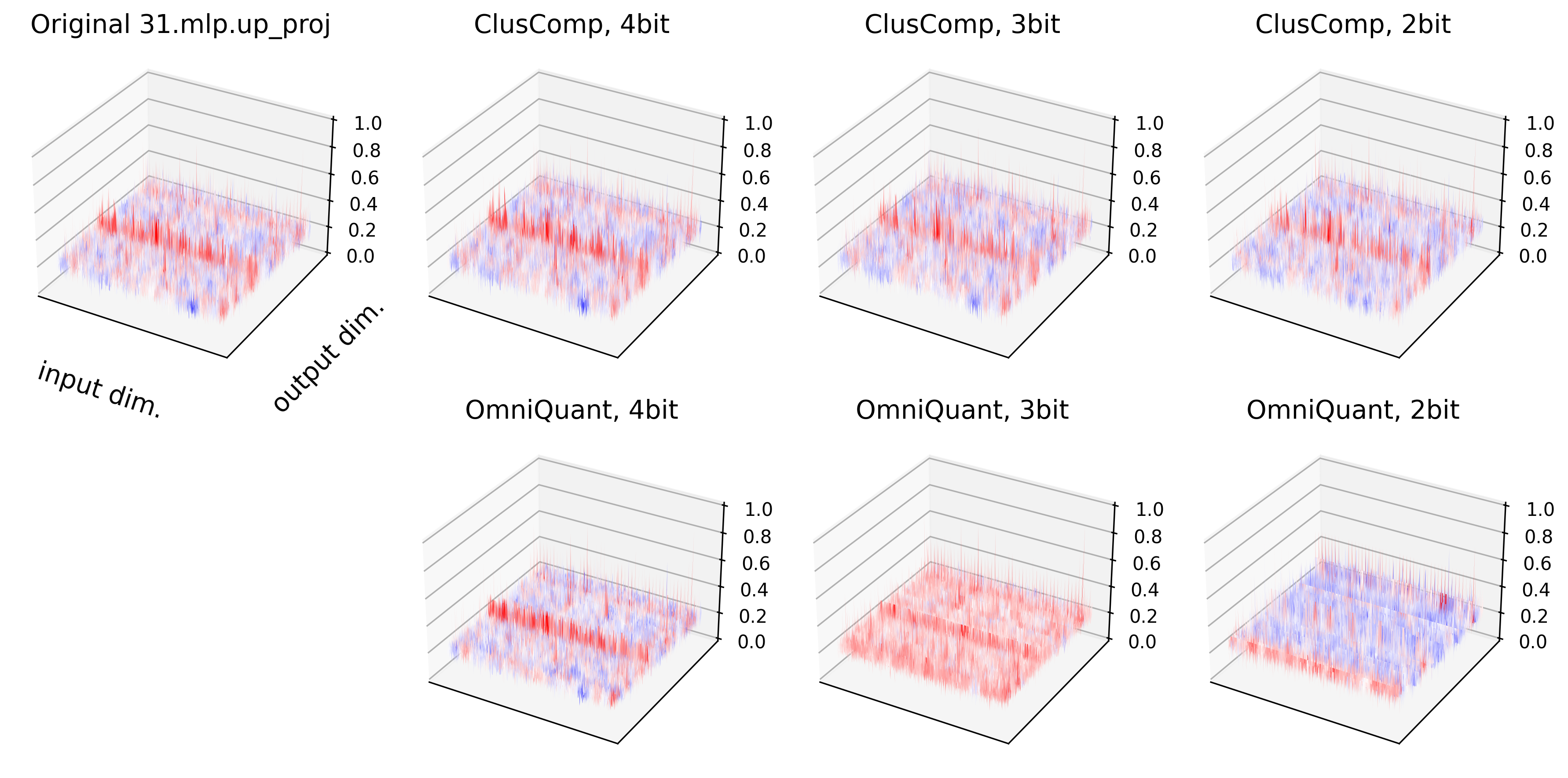}
     \caption{Up projection layer.}
 \end{subfigure}
 \begin{subfigure}[t]{0.48\textwidth}
     \centering
     \includegraphics[width=\textwidth]{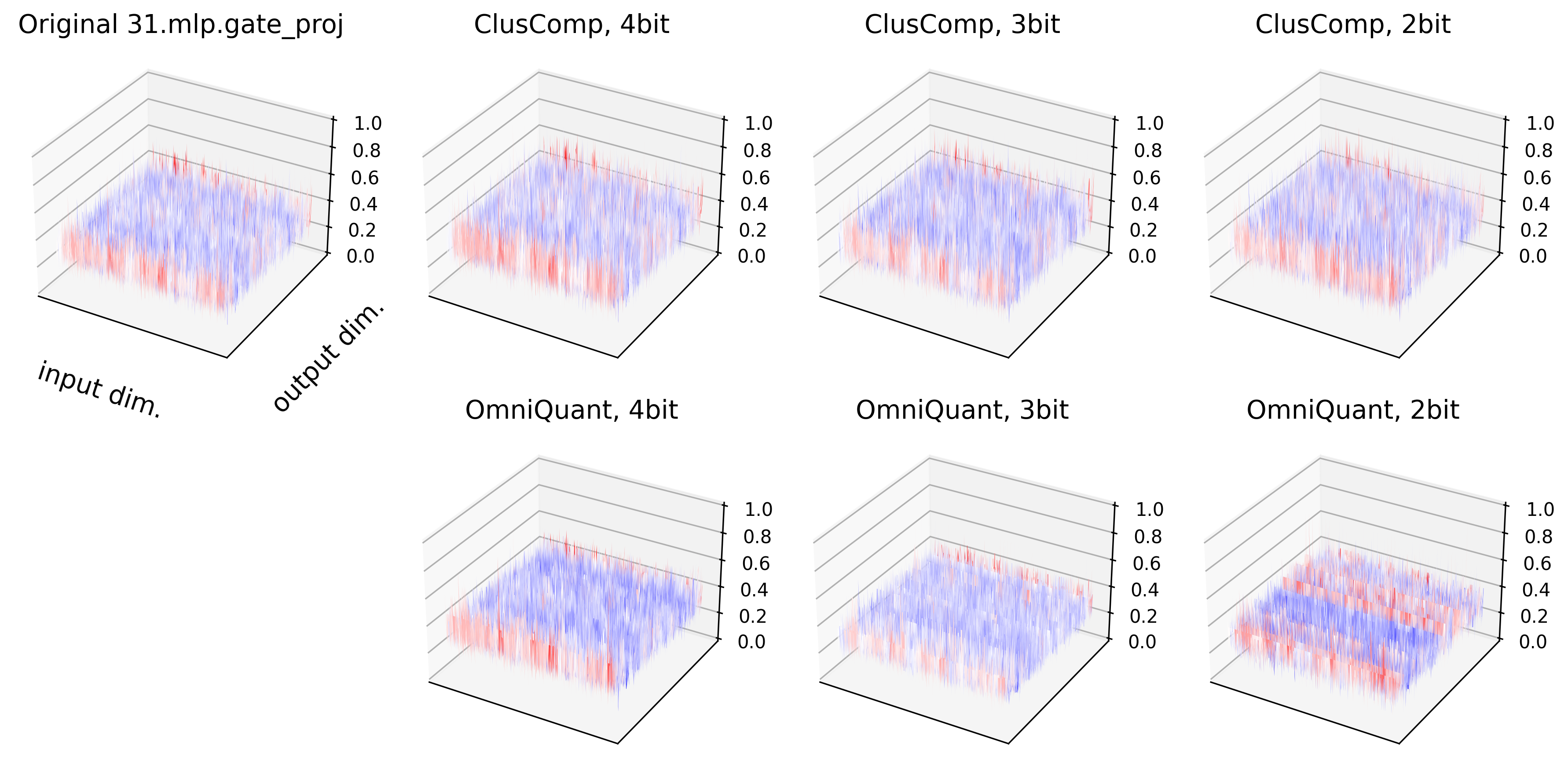}
     \caption{Gate projection layer.}
 \end{subfigure}
 \begin{subfigure}[t]{0.48\textwidth}
     \includegraphics[width=\textwidth]{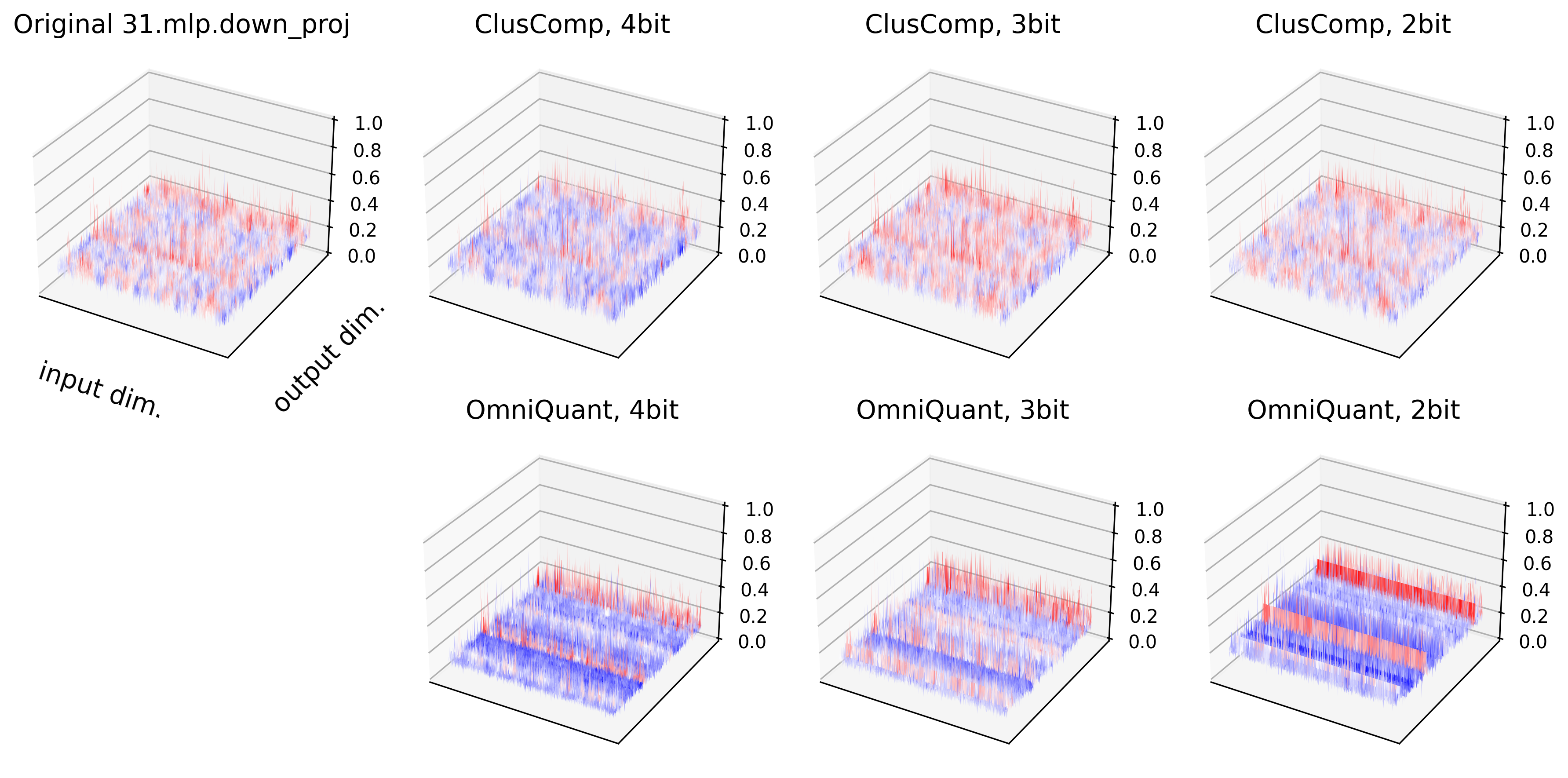}
     \caption{Down projection layer.}
 \end{subfigure}
    \caption[]{Weight patterns of all layers in the last block (31-st block) of Llama-2-7B. Darker red and blue indicate larger and smaller weight values, respectively. ClusComp's weight distribution of different bit levels can better simulate the original weight distribution.}
    \label{fig: last block}
\end{figure*}

To make the weight pattern more obvious, we apply these sequential processing steps: (1) take the absolute weight values; (2) downsample the grids with 8 $\times$ 8 maxpool kernels; (3) calculate the logarithm of these values; (4) normalize the log-values. We offer the visualization of all layers in the first and last blocks of Llama-2-7B in Figure \ref{fig: first block} and \ref{fig: last block}.

\subsection{More finetuning results}
\label{subsec: more finetuning results}
\textbf{In-domain finetuning.} We finetune Llama-2-7B on the training sets of WikiText2 and GSM8K \citep{gsm8k}, and report the perplexity and accuracy on their respective validation/test set. ClusComp is compared with two LoRA-based techniques: QLoRA \citep{qlora} and LoftQ \citep{loftq}. As shown in Table \ref{tab: wiki gsm8k}, ClusComp consistently achieves superior results with fewer bits (except at the 4-bit level on WikiText2, where it performs comparably to LoftQ), even outperforming LoRA-finetuning of the FP16 model on GSM8K with a 2.54-bit compressed LLM.

\textbf{General-domain finetuning.} We finetune the compressed LLMs on Alpaca-GPT3.5 \citep{alpaca} and evaluate them using the MMLU benchmark \citep{mmlu}. ClusComp is compared against baseline methods that merge trained LoRA modules into the quantized linear layers after finetuning, i.e. GPTQ-LoRA, QA-LoRA \citep{qalora} and PEQA \citep{peqa}.

As shown in Table \ref{tab: mmlu}, ClusComp consistently outperforms the baseline methods across different LLMs and bit-widths, while using fewer bits. The only exception occurs at the 4-bit level for Llama-1-7B, where ClusComp underperforms QA-LoRA by a small margin of 0.3 accuracy. The performance gap between ClusComp and baselines is enlarged for lower bits or recent LLM series.

\subsection{More baselines}
\label{subsec: more baselines}
In this section, we compare ClusComp against SqueezeLLM \citep{squeezellm} and AdaDim \citep{adadim}. As presented in Table \ref{tab: squeezellm}, ClusComp consistently achieves lower perplexity than SqueezeLLM, at comparable or even lower bit precision. Similarly, as shown in Table \ref{tab: adadim}, ClusComp outperforms AdaDim on both MMLU and CSR benchmarks.

\subsection{Compression performance of the multimodal backbone}
In Table \ref{tab: multimodal ppl}, we show the performance of the compressed Llama-3-8B backbone in LLaVA-Next-8B. ClusComp again shows its superior performance.

\begin{table*}[t]
    \begin{center}
    \scriptsize
    \begin{tabular}{llrllllll}
    \toprule
    \textbf{Method} & \textbf{Setting} & \textbf{\#Bit} & \textbf{1-7B} & \textbf{2-7B} & \textbf{2-13B} & \textbf{2-70B} & \textbf{3-8B} & \textbf{3-70B} \\
    \midrule
    - & - & 16.00 & 5.68 & 5.47 & 4.88 & 3.31 & 6.12 & 2.90 \\
    \cmidrule(lr){1-9}
    RTN & w4g128 & 4.13 & 5.96 & 5.72 & 4.98 & 3.46 & 8.50 & 3.60 \\
    GPTQ & w4g128 & 4.13 & 5.85 & 5.61 & 4.98 & 3.42 & 6.50 & 3.30 \\
    AWQ & w4g128 & 4.13 & 5.81 & 5.62 & 4.97 & - & 6.60 & 3.30 \\
    GPTVQ & w4g128 & 4.13 & - & 5.68 & 4.97 & \textbf{3.39} & - & - \\
    AffineQuant & w4g128 & 4.13 & 5.77 & 5.58 & 4.95 & - & - & - \\
    OmniQuant & w4g128 & 4.16 & 5.77 & 5.58 & 4.95 & 3.40 & - & - \\
    \rowcolor{green!8}
    \textcolor{gray!80}{ClusComp$^{-}$} & \textcolor{gray!80}{g4n65500} & \textcolor{gray!80}{$\leq$4.14} & \textcolor{gray!80}{5.88 (4.14)} & \textcolor{gray!80}{5.67 (4.14)} & \textcolor{gray!80}{5.04 (4.09)} & \textcolor{gray!80}{3.44 (4.03)} & \textcolor{gray!80}{6.59 (4.13)} & \textcolor{gray!80}{3.28 (4.03)} \\
    \rowcolor{green!8}
    \textbf{ClusComp} & g4n65500 & $\leq$4.14 & \textbf{5.73} (4.14) & \textbf{5.54} (4.14) & \textbf{4.94} (4.09) & 3.40 (4.03) & \textbf{6.39} (4.13) & \textbf{3.12} (4.03) \\
    \cmidrule(lr){1-9}
    RTN & w4 & 4.00 & 6.43 & 6.11 & 5.20 & 3.67 & 8.70 & - \\
    GPTQ & w4 & 4.00 & 6.13 & 5.83 & 5.13 & 3.58 & \textbf{7.00} & - \\
    AWQ & w4 & 4.00 & 6.08 & 6.15 & 5.12 & - & 7.10 & - \\
    QuIP & w4 & 4.00 & - & - & - & 3.53 & - & - \\
    AffineQuant & w4 & 4.00 & \textbf{5.84} & 5.69 & \textbf{5.01} & - & - & - \\
    OmniQuant & w4 & 4.00 & 5.86 & 5.74 & 5.02 & \textbf{3.47} & - & - \\
    \rowcolor{green!8}
    \textcolor{gray!80}{ClusComp$^{-}$} & \textcolor{gray!80}{g5n65500} & \textcolor{gray!80}{\textbf{3.38}} & \textcolor{gray!80}{6.27} & \textcolor{gray!80}{5.90} & \textcolor{gray!80}{-} & \textcolor{gray!80}{-} & \textcolor{gray!80}{-} & \textcolor{gray!80}{-} \\
    \rowcolor{green!8}
    \textbf{ClusComp} & g5n65500 & \textbf{3.38} & \textbf{5.84} & \textbf{5.67} & - & - & - & - \\    
    \cmidrule(lr){1-9}
    RTN & w3g128 & 3.13 & 7.01 & 6.66 & 5.51 & 3.97 & 27.91 & 11.84 \\
    GPTQ & w3g128 & 3.13 & 6.55 & 6.29 & 5.42 & 3.85 & 8.22 & 5.22 \\
    AWQ & w3g128 & 3.13 & 6.46 & 6.24 & 5.32 & - & \textbf{8.19} & \textbf{4.81} \\
    SliM-LLM$^{+}$ & w3g128 & 3.13 & \textbf{6.07} & 5.94 & 5.11 & \textbf{3.35} & - & - \\
    AffineQuant & w3g128 & 3.13 & 6.14 & 6.08 & 5.28 & - & - & - \\
    GPTVQ & w3g128 & 3.13 & - & \textbf{5.82} & \textbf{5.10} & 3.55 & - & - \\
    OmniQuant & w3g128 & 3.15 & 6.15 & 6.03 & 5.28 & 3.78 & - & - \\
    \cmidrule(lr){1-9}
    RTN & w3 & 3.00 & 25.73 & 5.4e2 & 10.68 & 7.52 & 2.2e3 & - \\
    GPTQ & w3 & 3.00 & 8.06 & 8.37 & 6.44 & 4.82 & 13.0 & - \\
    AWQ & w3 & 3.00 & 11.88 & 24.00 & 10.45 & - & 12.8 & - \\
    QuIP & w3 & 3.00 & - & - & - & 3.85 & 7.5 & - \\
    AffineQuant & w3 & 3.00 & 6.30 & 6.55 & 5.62 & - & - & - \\
    OmniQuant & w3 & 3.00 &  6.49 & 6.58 & 5.58 & 3.92 & - & - \\
    \rowcolor{green!8}
    \textcolor{gray!80}{ClusComp$^{-}$} & \textcolor{gray!80}{g6n65500} & \textcolor{gray!80}{$\leq$2.89} & \textcolor{gray!80}{6.74 (2.89)} & \textcolor{gray!80}{6.54 (2.89)} & \textcolor{gray!80}{6.27 (2.81)} & \textcolor{gray!80}{4.02 (2.72)} & \textcolor{gray!80}{8.77 (2.87)} & \textcolor{gray!80}{4.98 (2.72)} \\
    \rowcolor{green!8}
    \textbf{ClusComp} & g6n65500 & $\leq$\textbf{2.89} & \textbf{6.01} (2.89) & \textbf{5.86} (2.89) & \textbf{5.18} (2.81) & \textbf{3.72} (2.72) & \textbf{7.34} (2.87) & \textbf{4.63} (2.72) \\
    \cmidrule(lr){1-9}
    \rowcolor{green!8}
    \textcolor{gray!80}{ClusComp$^{-}$} & \textcolor{gray!80}{g7n65500} & \textcolor{gray!80}{2.54} & \textcolor{gray!80}{7.79} & \textcolor{gray!80}{7.64} & \textcolor{gray!80}{-} & \textcolor{gray!80}{-} & \textcolor{gray!80}{-} & \textcolor{gray!80}{-} \\
    \rowcolor{green!8}
    \textbf{ClusComp} & g7n65500 & 2.54 & \textbf{6.28} & \textbf{6.15} & - & - & - & - \\
    \cmidrule(lr){1-9}
    RTN & w2g64 & 2.25 & 1.9e2 & 4.3e2 & 26.22 & 10.31 & - & - \\
    GPTQ & w2g64 & 2.25 & 22.10 & 20.85 & 22.44 & NAN & 1.8e2 & - \\
    AWQ & w2g64 & 2.25 & 2.5e5 & 2.1e5 & 1.2e5 & - & - & - \\
    GPTVQ & w2g64 & 2.25 & - & 7.22 & 6.08 & \textbf{4.39} & - & - \\ 
    AffineQuant & w2g64 & 2.25 & 8.35 & 9.05 & 7.11 & - & - & - \\
    OmniQuant & w2g64 & 2.28 & 8.90 & 9.62 & 7.56 & 6.11 & - & - \\
    \rowcolor{green!8}
    \textcolor{gray!80}{\textbf{ClusComp}$^{-}$} & \textcolor{gray!80}{g8n65500} & \textcolor{gray!80}{$\leq$2.29} & \textcolor{gray!80}{10.25 (2.29)} & \textcolor{gray!80}{11.10 (2.29)} & \textcolor{gray!80}{14.39 (2.19)} & \textcolor{gray!80}{-} & \textcolor{gray!80}{29.20 (2.27)} & \textcolor{gray!80}{-} \\
    \rowcolor{green!8}
    \textbf{ClusComp} & g8n65500 & $\leq$2.29 & \textbf{6.66} (2.29) & \textbf{6.61} (2.29) & \textbf{5.74} (2.19) & - & \textbf{9.68} (2.27) & - \\
    \cmidrule(lr){1-9}
    RTN & w2g128 & 2.13 & 1.9e3 & 4.2e3 & 1.2e2 & 27.27 & 1.9e3 & 4.6e5 \\
    GPTQ & w2g128 & 2.13 & 44.01 & 36.77 & 28.14 & NAN & 2.1e2 & 11.9 \\
    AWQ & w2g128 & 2.13 & 2.6e5 & 2.2e5 & 1.2e5 & - & 1.7e6 & 1.7e6 \\
    SliM-LLM$^{+}$ & w2g128 & 2.13 & 9.68 & 10.87 & 7.59 & 6.44 & - & - \\
    QuIP & w2g128 & 2.13 & - & 39.73 & 13.48 & 6.64 & 84.97 & 13.03 \\
    PB-LLM & w2g128 & 2.13 & - & 25.37 & 49.81 & NAN & 44.12 & 11.68 \\
    GPTVQ & w2g128 & 2.13 & - & 8.23 & 6.50 & 4.64 & - & - \\
    AffineQuant & w2g128 & 2.13 & 13.51 & 10.87 & 7.64 & -  & - \\
    OmniQuant & w2g128 & 2.14 & 9.72 & 11.06 & 8.26 & 6.55 & - & - \\
    \rowcolor{green!8}
    \textcolor{gray!80}{ClusComp$^{-}$} & \textcolor{gray!80}{g8} & \textcolor{gray!80}{$\leq$2.15} & \textcolor{gray!80}{28.76} & \textcolor{gray!80}{21.9} & \textcolor{gray!80}{14.50} & \textcolor{gray!80}{5.43} & \textcolor{gray!80}{2.1e2} & \textcolor{gray!80}{11.40} \\
    \rowcolor{green!8}
    \textbf{ClusComp} & g8 & $\leq$2.15 & \textbf{7.06} & \textbf{7.04} & \textbf{5.85} & \textbf{4.37} & \textbf{11.57} & \textbf{7.61} \\
    \rowcolor{green!8}
    & & & (2.15,n35000) & (2.15,n35000) & (2.14,n50000) & (2.07,n65500) & (2.14,n35000) & (2.07,n65500) \\
    \cmidrule(lr){1-9}
    \rowcolor{green!8}
    \textcolor{gray!80}{ClusComp$^{-}$} & \textcolor{gray!80}{g9n65500} & \textcolor{gray!80}{2.11} & \textcolor{gray!80}{-} & \textcolor{gray!80}{22.71} & \textcolor{gray!80}{-} & \textcolor{gray!80}{-} & \textcolor{gray!80}{-} & \textcolor{gray!80}{-} \\
    \rowcolor{green!8}
    \textbf{ClusComp} & g9n65500 & 2.11 & - & \textbf{7.12} & - & - & - & - \\
    \cmidrule(lr){1-9}
    RTN & w2 & 2.00 & 1.1e5 & 3.8e4 & 5.6e4 & 2.0e4 & 2.7e6 & - \\
    GPTQ & w2 & 2.00 & 2.1e3 & 7.7e3 & 2.1e3 & 77.95 & 5.7e4 & - \\
    QuIP & w2 & 2.00 & - & - & - & 6.33 & 85.1 & - \\
    AffineQuant & w2 & 2.00 & 9.53 & 35.07 & 12.42 & - & - & - \\
    OmniQuant & w2 & 2.00 & 15.47 & 37.37 & 17.21 & 7.81 & - & - \\
    \rowcolor{green!8}
    \textcolor{gray!80}{ClusComp$^{-}$} & \textcolor{gray!80}{g9} & \textcolor{gray!80}{$\leq$2.01} & \textcolor{gray!80}{65.09} & \textcolor{gray!80}{52.38} & \textcolor{gray!80}{22.90} & \textcolor{gray!80}{9.84} & \textcolor{gray!80}{3.1e2} & \textcolor{gray!80}{-} \\
    \rowcolor{green!8}
    \textbf{ClusComp} & g9 & $\leq$2.01 & \textbf{7.49} & \textbf{7.50} & \textbf{6.17} & \textbf{4.83} & \textbf{12.33} & - \\
    \rowcolor{green!8}
    & & & (2.00,n45000) & (2.00,n45000) & (1.99,n65500) & (1.85,n65500) & (2.01,n50000) & - \\
    \bottomrule
    \end{tabular}
    \caption{The full perplexity results of Llama series on WikiText2. ``g'' and ``n'' denote the dimension and number of centroids in the codebook, respectively. The number in the brackets is the exact bits of different settings for different LLMs.}
    \label{tab: wiki ppl}
    \end{center}
\end{table*}

\begin{table*}[t]
    \begin{center}
    \scriptsize
    \begin{tabular}{llrllllll}
    \toprule
    \textbf{Method} & \textbf{Setting} & \textbf{\#Bit} & \textbf{1-7B} & \textbf{2-7B} & \textbf{2-13B} & \textbf{2-70B} & \textbf{3-8B} & \textbf{3-70B} \\
    \midrule
    - & - & 16.00 & 7.08 & 6.97 & 6.46 & 5.52 & 9.20 & 5.87  \\
    \cmidrule(lr){1-9}
    RTN & w4g128 & 4.13 & 7.37 & 7.24 & 6.58 & 5.63 & 13.40 & 8.90  \\
    GPTQ & w4g128 & 4.13 & 7.21 & 7.12 & 6.56 & \textbf{5.58} & 10.40 & \textbf{6.94}  \\
    AWQ & w4g128 & 4.13 & 7.21 & 7.13 & 6.56 & - & 9.40 & 7.00  \\
    AffineQuant & w4g128 & 4.13 &  7.20 & 7.12 & 6.56 & - & - & - \\
    OmniQuant & w4g128 & 4.16 &  7.21 & 7.12 & 6.56 & \textbf{5.58} & - & - \\
    \rowcolor{green!8}
    \textcolor{gray!80}{ClusComp$^{-}$} & \textcolor{gray!80}{g4n65500} & \textcolor{gray!80}{$\leq$4.14} & \textcolor{gray!80}{7.27 (4.14)} & \textcolor{gray!80}{7.16 (4.14)} & \textcolor{gray!80}{6.63 (4.09)} & \textcolor{gray!80}{5.61 (4.03)} & \textcolor{gray!80}{9.39 (4.13)} & \textcolor{gray!80}{7.02 (4.03)} \\
    \rowcolor{green!8}
    \textbf{ClusComp} & g4n65500 & $\leq$4.13 & \textbf{7.17} (4.14) & \textbf{7.09} (4.14) & \textbf{6.55} (4.09) & 5.61 (4.03) & \textbf{9.27} (4.13) & 6.99 (4.03) \\   
    \cmidrule(lr){1-9}
    RTN & w3 & 3.00 & 28.26 & 4.0e2 & 12.51 & 10.02 & 2.2e3 & - \\
    GPTQ & w3 & 3.00 & 9.49 & 9.81 & 8.02 & 6.57 & 13.0 & - \\
    AWQ & w3 & 3.00 & 13.26 & 23.85 & 13.07 & - & 12.8 & - \\
    QuIP & w3 & 3.00 & - & - & - & 6.14 & - & - \\
    AffineQuant & w3 & 3.00 & 8.03 & 8.57 & 7.56 & - & - & - \\
    OmniQuant & w3 & 3.00 & 8.19 & 8.65 & 7.44 & 6.06 & - & -  \\
    \rowcolor{green!8}
    \textcolor{gray!80}{ClusComp$^{-}$} & \textcolor{gray!80}{g6n65500} & \textcolor{gray!80}{$\leq$2.89} & \textcolor{gray!80}{8.14 (2.89)} & \textcolor{gray!80}{8.19 (2.89)} & \textcolor{gray!80}{8.21 (2.81)} & \textcolor{gray!80}{6.06 (2.72)} & \textcolor{gray!80}{12.41 (2.87)} & \textcolor{gray!80}{8.26 (2.72)} \\
    \rowcolor{green!8}
    \textbf{ClusComp} & g6n65500 & $\leq$\textbf{2.89} & \textbf{7.64} (2.89) & \textbf{7.61} (2.89) & \textbf{6.91} (2.81) & \textbf{5.86} (2.72) & \textbf{11.31} (2.87) & \textbf{8.26} (2.72) \\
    \midrule
    \rowcolor{green!8}
    \textcolor{gray!80}{ClusComp$^{-}$} & \textcolor{gray!80}{g7n65500} & \textcolor{gray!80}{2.54} & \textcolor{gray!80}{9.46} & \textcolor{gray!80}{9.51} & \textcolor{gray!80}{-} & \textcolor{gray!80}{-} & \textcolor{gray!80}{-} & \textcolor{gray!80}{-} \\
    \rowcolor{green!8}
    \textbf{ClusComp} & g7n65500 & 2.54 & \textbf{8.10} & \textbf{8.13} & - & - & - & - \\
    \cmidrule(lr){1-9}
    RTN & w2g64 & 2.25 & 1.5e2 & 4.8e2 & 28.69 & 13.43 & - & - \\
    GPTQ & w2g64 & 2.25 & 17.71 & 19.40 & 12.48 & NAN & - & - \\
    AWQ & w2g64 & 2.25 & 2.8e5 & 1.6e5 & 9.5e4 & - & - & - \\
    OmniQuant & w2g64 & 2.28 & 11.78 & 12.72 & 10.05 & 7.88 & - & - \\
    \rowcolor{green!8}
    \textcolor{gray!80}{ClusComp$^{-}$} & \textcolor{gray!80}{g8n65500} & \textcolor{gray!80}{$\leq$2.29} & \textcolor{gray!80}{13.06 (2.29)} & \textcolor{gray!80}{14.07 (2.29)} & \textcolor{gray!80}{19.75 (2.19)} & \textcolor{gray!80}{-} & \textcolor{gray!80}{38.68 (2.27)} & \textcolor{gray!80}{-} \\
    \rowcolor{green!8}
    \textbf{ClusComp} & g8n65500 & $\leq$2.29 & \textbf{8.76} (2.29) & \textbf{8.88} (2.29) & \textbf{7.75} (2.19) & - & \textbf{15.57} (2.27) & - \\
    \cmidrule(lr){1-9}
    RTN & w2g128 & 2.13 &  1.0e3 & 4.9e3 & 1.4e2 & 42.13 & 1.9e3 & - \\
    GPTQ & w2g128 & 2.13 & 27.71 & 33.70 & 20.97 & NAN & 2.1e2 & - \\
    AWQ & w2g128 & 2.13 & 1.9e5 & 1.7e5 & 9.4e4 & - & 1.7e6 & -  \\
    SliM-LLM$^{+}$ & w2g128 & 2.13 & 14.99 & 18.18 & 10.24 & 8.40 & - & - \\
    QuIP & w2g128 & 2.13 & - & 31.94 & 16.16 & 8.17 & 1.3e2 & 22.24  \\
    PB-LLM & w2g128 & 2.13 & - & 29.84 & 19.82 & 8.95 & 79.21 & 33.91 \\
    AffineQuant & w2g128 & 2.13 & - & 16.02 & 10.98 & - & - & - \\
    OmniQuant & w2g128 & 2.14 & 12.97 & 15.02 & 11.05 & 8.52 & - & - \\
    \rowcolor{green!8}
    \textcolor{gray!80}{ClusComp$^{-}$} & \textcolor{gray!80}{g8} & \textcolor{gray!80}{$\leq$2.15} & \textcolor{gray!80}{29.67} & \textcolor{gray!80}{25.26} & \textcolor{gray!80}{18.83} & \textcolor{gray!80}{7.59} & \textcolor{gray!80}{1.9e2} & \textcolor{gray!80}{16.52} \\
    \rowcolor{green!8}
    \textbf{ClusComp} & g8 & $\leq$2.15 & \textbf{9.33} & \textbf{9.49} & \textbf{7.92} & \textbf{6.44} & \textbf{17.89} & \textbf{10.81} \\
    \rowcolor{green!8}
    & & & (2.15,n35000) & (2.15,n35000) & (2.14,n50000) & (2.07,n65500) & (2.14,n35000) & (2.07,n65500) \\
    \cmidrule(lr){1-9}
    \rowcolor{green!8}
    \textcolor{gray!80}{ClusComp$^{-}$} & \textcolor{gray!80}{g9n65500} & \textcolor{gray!80}{2.11} & \textcolor{gray!80}{-} & \textcolor{gray!80}{27.37} & \textcolor{gray!80}{-} & \textcolor{gray!80}{-} & \textcolor{gray!80}{-} & \textcolor{gray!80}{-} \\
    \rowcolor{green!8}
    \textbf{ClusComp} & g9n65500 & 2.11 & - & \textbf{9.73} & - & - & - & - \\
    \midrule
    RTN & w2 & 2.00 & 1.3e5 & 4.8e4 & 7.2e4 & 2.4e4 & 2.7e6 & - \\
    GPTQ & w2 & 2.00 & 6.9e2 & NAN & 3.2e2 & 48.82 & 5.7e4 & - \\
    QuIP & w2 & 2.00 & - & - & - & - & 1.3e2 & -\\
    OmniQuant & w2 & 2.00 &  24.89 & 90.64 & 26.76 & 12.28 & 8.2e5 \\
    \rowcolor{green!8}
    \textcolor{gray!80}{ClusComp$^{-}$} & \textcolor{gray!80}{g9} & \textcolor{gray!80}{$\leq$2.01} & \textcolor{gray!80}{74.61} & \textcolor{gray!80}{50.08} & \textcolor{gray!80}{24.47} & \textcolor{gray!80}{13.96} & \textcolor{gray!80}{2.2e2} & \textcolor{gray!80}{-}  \\
    \rowcolor{green!8}
    \textbf{ClusComp} & g9 & $\leq$2.01 & \textbf{10.11} & \textbf{10.29} & \textbf{8.49} & \textbf{7.02} & \textbf{21.45} & - \\
    \rowcolor{green!8}
    & & & (2.00,n45000) & (2.00,n45000) & (1.99,n65500) & (1.85,n65500) & (2.01,n50000) & - \\
    \bottomrule
    \end{tabular}
    \caption{The full perplexity results of Llama series on C4. ``g'' and ``n'' denote the dimension and number of centroids in the codebook, respectively. The number in the brackets is the exact bits of different settings for different LLMs.}
    \label{tab: c4 ppl}
    \end{center}
\end{table*}

\begin{table*}[t]
    \begin{center}
    \scriptsize
    \begin{tabular}{lrccccccccccc}
    \toprule
    \multirow{3}{*}{\textbf{Method}} & \multirow{3}{*}{\textbf{\#Bit}} & \multicolumn{2}{c}{\textbf{PIQA}} & \multicolumn{2}{c}{\textbf{ARC-e}} & \multicolumn{2}{c}{\textbf{ARC-c}} & \textbf{BoolQ} & \multicolumn{2}{c}{\textbf{HellaSwag}} & \textbf{WinoGrande} & \multirow{3}{*}{\textbf{Avg}} \\
    \cmidrule(lr){3-4} \cmidrule(lr){5-6} \cmidrule(lr){7-8} \cmidrule(lr){10-11}
     & & acc & \textbf{acc\_n} & acc & \textbf{acc\_n} & acc & \textbf{acc\_n} & \textbf{acc} & acc & \textbf{acc\_n} & \textbf{acc} &  \\
    \hline
    \midrule
     \textbf{Llama-2-7B} & 16.00 & - & 79.1 & - & 74.6 & - & 46.3 & 77.7 & - & 76.0 & 69.1 & 70.5 \\
     \cmidrule(lr){1-13}
     \rowcolor{green!8}
     \textbf{ClusComp} & 4.14 & 77.5 & 79.2 & 75.3 & 72.8 & 42.7 & 45.3 & 76.2 & 56.5 & 75.1 & 68.9 & 69.6 \\
     \cmidrule(lr){1-13}
     RTN & 3.13 & - & 76.8 & - & 70.5 & - & 42.9 & 71.7 & - & \textbf{74.0} & 67.6 & 67.3 \\
     GPTQ & 3.13 & - & 77.4 & - & 68.1 & - & 40.7 & 71.0 & - & 72.5 & 67.3 & 66.2 \\
     GPTVQ & 3.13 & - & \textbf{77.6} & - & \textbf{72.7} & - & \textbf{43.7} & 71.7 & - & 72.7 & 67.6 & 67.7 \\
     \rowcolor{green!8}
     \textbf{ClusComp} & \textbf{2.89} & 76.8 & \textbf{77.6} & 74.4 & 71.3 & 42.3 & 42.9 & \textbf{74.6} & 54.4 & 72.4 & \textbf{68.8} & \textbf{67.9} \\
     \cmidrule(lr){1-13}
     RTN & 2.25 & - & 58.8 & - & 36.7 & - & 24.8 & 41.9 & - & 40.4 & 51.9 & 42.4 \\
     GPTQ & 2.25 & - & 60.8 & - & 39.0 & - & 25.2 & 59.3 & - & 45.8 & 55.5 & 47.6 \\
     GPTVQ & 2.25 & - & 73.3 & - & 63.4 & - & 35.9 & 66.3 & - & 63.9 & \textbf{66.1} & 61.5 \\
     \rowcolor{green!8}
     \textbf{ClusComp} & 2.29 & 74.9 & \textbf{76.0} & 69.8 & \textbf{65.2} & 37.7 & \textbf{37.4} & \textbf{73.0} & 51.1 & \textbf{68.4} & 65.0 & \textbf{64.1} \\
     \cmidrule(lr){1-13}
     \rowcolor{green!8}
     \textbf{ClusComp} & 2.15 & 74.3 & 75.1 & 69.6 & 65.2 & 35.7 & 38.4 & 69.5 & 49.2 & 66.4 & 63.5 & 63.0 \\
     \cmidrule(lr){1-13}
     RTN & 2.13 & - & 51.1 & - & 28.0 & - & 25.0 & 41.1 & - & 26.6 & 49.9 & 36.9 \\
     GPTQ & 2.13 & - & 54.8 & - & 30.6 & - & 25.1 & 53.4 & - & 33.1 & 51.5 & 41.4 \\
     GPTVQ & 2.13 & - & 70.7 & - & 58.1 & - & 31.5 & 63.7 & - & 58.5 & 60.9 & 57.2 \\
     \rowcolor{green!8}
     \textbf{ClusComp} & \textbf{2.00} & 72.6 & \textbf{73.7} & 67.0 & \textbf{62.8} & 32.9 & \textbf{36.6} & \textbf{70.9} & 47.2 & \textbf{63.5} & \textbf{63.4} & \textbf{61.8} \\
     \hline
     \midrule
     \textbf{Llama-2-13B} & 16.00 &  - & 80.5 & - & 77.5 & - & 49.2 & 80.5 & - & 79.4 & 72.1 & 73.2 \\
     \cmidrule(lr){1-13}
     \rowcolor{green!8}
     \textbf{ClusComp} & 4.09 & 78.9 & 79.9 & 78.9 & 76.9 & 47.7 & 49.2 & 81.4 & 60.0 & 79.0 & 72.4 & 73.1 \\
     \cmidrule(lr){1-13}
     RTN & 3.13 & - & 78.9 & - & 74.3 & - & 46.8 & 77.3 & - & 76.5 & 70.8 & 70.8 \\
     GPTQ & 3.13 & - & 79.3 & - & 75.8 & - & 47.0 & 78.9 & - & \textbf{77.2} & 70.4 & 71.4 \\
     GPTVQ & 3.13 & - & 79.4 & - & 75.3 & - & \textbf{48.1} & 79.0 & - & 77.0 & \textbf{71.7} & 71.8 \\
     \rowcolor{green!8}
     \textbf{ClusComp} & \textbf{2.81} & 78.7 & \textbf{79.7} & 78.5 & \textbf{76.7} & 45.9 & 47.8 & \textbf{80.7} & 58.3 & 76.8 & 71.4 & \textbf{72.2} \\
     \cmidrule(lr){1-13}
     RTN & 2.25 & - & 61.6 & - & 41.6 & - & 25.4 & 49.8 & - & 48.2 & 51.9 & 46.4 \\
     GPTQ & 2.25 & - & 70.1 & - & 56.7 & - & 31.6 & 51.1 & - & 56.6 & 58.9 & 54.2 \\
     GPTVQ & 2.25 & - & 76.2 & - & 71.9 & - & 43.3 & 67.6 & - & 70.0 & 68.2 & 66.2 \\
     \rowcolor{green!8}
     \textbf{ClusComp} & 2.19 & 76.6 & \textbf{77.3} & 75.0 & \textbf{72.9} & 40.8 & \textbf{43.9} & \textbf{78.1} & 55.3 & \textbf{73.3} & \textbf{68.4} & \textbf{69.0} \\
     \cmidrule(lr){1-13}
     \rowcolor{green!8}
     \textbf{ClusComp} & 2.14 & 76.7 & 77.1 & 73.5 & 71.6 & 39.9 & 42.8 & 77.5 & 54.6 & 73.1 & 68.0 & 68.4 \\
     \cmidrule(lr){1-13}
     RTN & 2.13 & - & 58.4 & - & 32.3 & - & 25.5 & 47.9 & - & 39.4 & 48.9 & 42.1 \\
     GPTQ & 2.13 & - & 59.5 & - & 40.2 & - & 27.7 & 57.1 & - & 41.6 & 53.4 & 46.6 \\
     GPTVQ & 2.13 & - & 75.2 & - & 68.3 & - & 39.5 & 70.7 & - & 65.7 & \textbf{67.5} & 64.5 \\
     \rowcolor{green!8}
     \textbf{ClusComp} & \textbf{1.99} & 75.6 & \textbf{77.7} & 74.7 & \textbf{73.6} & 39.9 & \textbf{42.1} & \textbf{74.0} & 53.0 & \textbf{71.0} & 67.1 & \textbf{67.6} \\
    \bottomrule
    \end{tabular}
    \caption{Zero-shot evaluation of the quantized Llama-2-7B and Llama-2-13B, with baseline results taken from \citet{gptvq}. ``acc'' and ``acc\_n'' mean accuracy and normalized accuracy, respectively. We offer the results of all metrics for a convenient comparison of the follow-up works. But only the highlighted \textbf{metrics} are used to calculate the average accuracy.}
    \label{tab: commonsense 1}
    \end{center}
\end{table*}

\begin{table*}[t]
    \begin{center}
    \scriptsize
    \begin{tabular}{lrccccccccccc}
    \toprule
     \multirow{3}{*}{\textbf{Method}} &  \multirow{3}{*}{\textbf{\#Bit}} & \multicolumn{2}{c}{\textbf{PIQA}} & \multicolumn{2}{c}{\textbf{ARC-e}} & \multicolumn{2}{c}{\textbf{ARC-c}} & \textbf{BoolQ} & \multicolumn{2}{c}{\textbf{HellaSwag}} & \textbf{WinoGrande} & \multirow{3}{*}{\textbf{Avg}} \\
     \cmidrule(lr){3-4} \cmidrule(lr){5-6} \cmidrule(lr){7-8} \cmidrule(lr){10-11}
     & & \textbf{acc} & acc\_n & \textbf{acc} & acc\_n & \textbf{acc} & acc\_n & acc & \textbf{acc} & acc\_n & \textbf{acc} &  \\
     \midrule
     \textbf{Llama-3-8B} & 16.00 & 79.9 & - & 80.1 & - & 50.4 & - & - & 60.2 & - & 72.8 & 68.6 \\
     \cmidrule(lr){1-13}
     RTN & 4.13 & 76.6 & - & 70.1 & - & 45.0 & - & - & 56.8 & - & 71.0 & 63.9 \\
     GPTQ & 4.13 & 78.4 & - & 78.8 & - & 47.7 & - & - & 59.0 & - & 72.6 & 67.3 \\
     AWQ & 4.13 & \textbf{79.1} & - & 79.7 & - & 49.3 & - & - & 59.1 & - & \textbf{74.0} & 68.2 \\
     SliM-LLM & 4.13 & 78.9 & - & 79.9 & - & 49.4 & - & - & 58.7 & - & 72.6 & 67.9 \\
     \rowcolor{green!8}
     \textbf{ClusComp} & 4.13 & \textbf{79.1} & 80.5 & \textbf{80.9} & 79.6 & \textbf{49.7} & 54.1 & 81.1 & \textbf{59.3} & 78.3 & 72.9 & \textbf{68.4} \\
     \cmidrule(lr){1-13}
     RTN & 3.13 & 62.3 & - & 32.1 & - & 22.5 & - & - & 29.1 & - & 54.7 & 40.2 \\
     GPTQ & 3.13 & 74.9 & - & 70.5 & - & 37.7 & - & - & 54.3 & - & 71.1 & 61.7 \\
     AWQ & 3.13 & 77.7 & - & 74.0 & - & 43.2 & - & - & 55.1 & - & 72.1 & 64.4 \\
     SliM-LLM & 3.13 & \textbf{77.8} & - & 73.7 & - & 42.9 & - & - & 55.5 & - & \textbf{72.8} & 64.5 \\
     RTN & 3.00 & 56.2 & - & 31.1 & - & 20.0 & - & - & 27.5 & - & 53.1 & 35.6 \\
     GPTQ & 3.00 & 60.8 & - & 38.8 & - & 22.3 & - & - & 41.8 & - & 60.9 & 44.9 \\
     AWQ & 3.00 & 71.9 & - & 66.7 & - & 35.1 & - & - & 50.7 & - & 64.7 & 57.8 \\
     QuIP & 3.00 & 76.8 & - & 72.9 & - & 41.0 & - & - & 55.4 & - & 72.5 & 63.7 \\
     \rowcolor{green!8}
     \textbf{ClusComp} & \textbf{2.87} & 77.7 & 78.8 & \textbf{76.0} & 74.5 & \textbf{43.9} & 47.6 & 79.0 & \textbf{56.0} & 74.6 & 71.0 & \textbf{64.9} \\
     \cmidrule(lr){1-13}
     \rowcolor{green!8}
     \textbf{ClusComp} & 2.27 & 70.6 & 71.8 & 63.5 & 57.4 & 31.4 & 35.5 & 74.7 & 49.6 & 66.1 & 67.1 & 56.4 \\
     \cmidrule(lr){1-13}
     RTN & 2.13 & 53.1 & - & 24.8 & - & 22.1 & - & - & 26.9 & - & 53.1 & 36.0 \\
     GPTQ & 2.13 & 53.9 & - & 28.8 & - & 19.9 & - & - & 27.7 & - & 50.5 & 36.2 \\
     AWQ & 2.13 & 52.4 & - & 24.2 & - & 21.5 & - & - & 25.6 & - & 50.7 & 34.9 \\
     SliM-LLM & 2.13 & 57.1 & - & 35.4 & - & 26.1 & - & - & 28.9 & - & 56.6 & 40.8 \\
     PB-LLM & 2.13 & 57.0 & - & 37.8 & - & 17.2 & - & - & 29.8 & - & 52.5 & 38.8 \\
     \rowcolor{green!8}
     \textbf{ClusComp} & 2.14 & \textbf{68.0} & 67.1 & \textbf{54.7} & 49.0 & \textbf{26.4} & 28.8 & 71.5 & \textbf{47.0} & 63.0 & \textbf{62.4} & \textbf{51.7} \\
     \cmidrule(lr){1-13}
     RTN & 2.00 & 53.1 & - & 24.7 & - & 21.9 & - & - & 25.6 & - & 51.1 & 35.3 \\
     GPTQ & 2.00 & 52.8 & - & 25.0 & - & 20.5 & - & - & 26.6 & - & 49.6 & 34.9 \\
     AWQ & 2.00 & 55.2 & - & 25.2 & - & 21.3 & - & - & 25.4 & - & 50.4 & 35.5 \\
     QuIP & 2.00 & 52.9 & - & 29.0 & - & 21.3 & - & - & 29.2 & - & 51.7 & 36.8 \\
     \rowcolor{green!8}
     \textbf{ClusComp} & 2.01 & \textbf{70.1} & 69.6 & \textbf{63.3} & 57.7 & \textbf{31.9} & 34.2 & 66.6 & \textbf{44.4} & 58.0 & \textbf{58.4} & \textbf{53.6} \\
    \bottomrule
    \end{tabular}
    \caption{Zero-shot evaluation of the quantized Llama-3-8B, with baseline results taken from \citep{llama3quant}. ``acc'' and ``acc\_n'' mean accuracy and normalized accuracy, respectively. We offer the results of all metrics for a convenient comparison of the follow-up works. But only the highlighted \textbf{metrics} (excluding BoolQ) are used to calculate the average accuracy.}
    \label{tab: commonsense 2}
    \end{center}
\end{table*}

\begin{table*}[t]
    \begin{center}
    \footnotesize
    \begin{tabular}{lrcccccc}
    \toprule
    \textbf{Method} & \textbf{\#Bit} & \textbf{PIQA} & \textbf{ArcE} & \textbf{ArcC} & \textbf{Hella.} & \textbf{Wino.} &\textbf{Avg} \\
    \midrule
    \textbf{Llama-2-7B} & 16.00  & 78.1 & 76.3 & 43.4 & 57.1 & 69.1 & 64.8 \\
    QuIP\# & 2.02 & \textbf{75.1} & 64.6 & \textbf{34.6} & 48.3 & 64.9 & 57.5 \\
    AQLM & 2.02 & 73.6 & 61.9 & 33.3 & \textbf{49.5} & 64.2 & 56.5 \\
    \rowcolor{green!8}
    \textcolor{gray!80}{ClusComp} & \textcolor{gray!80}{2.00} & \textcolor{gray!80}{72.6} & \textcolor{gray!80}{67.0} & \textcolor{gray!80}{32.9} & \textcolor{gray!80}{47.2} & \textcolor{gray!80}{63.4} & \textcolor{gray!80}{56.6} \\
    \rowcolor{green!8}
    \textbf{ClusComp$^+$} & 2.00 & 73.5 & \textbf{67.2} & 33.9 & 49.3 & \textbf{65.1} & \textbf{57.8} \\
    \midrule
    \textbf{Llama-2-13B} & 16.00 & 79.1 & 79.4 & 48.4 & 60.0 & 72.2 & 67.8 \\
    QuIP\# & 2.01 & \textbf{77.3} & 69.3 & 39.5 & 53.4 & 67.7 & 61.5 \\
    AQLM & 1.97 & 76.2 & 69.8 & 37.8 & 53.7 & 65.4 & 60.6 \\
    \rowcolor{green!8}
    \textcolor{gray!80}{ClusComp} & \textcolor{gray!80}{1.99} & \textcolor{gray!80}{75.6} & \textcolor{gray!80}{74.7} & \textcolor{gray!80}{39.9} & \textcolor{gray!80}{53.0} & \textcolor{gray!80}{67.1} & \textcolor{gray!80}{62.0} \\
    \rowcolor{green!8}
    \textbf{ClusComp$^+$} & 1.99 & 76.8 & \textbf{74.9} & \textbf{40.7} & \textbf{54.5} & \textbf{68.4} & \textbf{63.1} \\
    \midrule
    \textbf{Llama-3-8B} & 16.00 & 79.7 & 80.1 & 50.4 & 60.2 & 72.6 & 68.6 \\
    QuiP & 2.00 & 52.9 & 29.0 & 21.3 & 29.2 & 51.7 & 36.8 \\
    PB-LLM & 2.00 & 57.0 & 37.8 & 17.2 & 29.8 & 52.5 & 38.8 \\
    DB-LLM & 2.00 & 68.9 & 59.1 & 28.2 & 42.1 & 60.4 & 51.8 \\
    \rowcolor{green!8}
    \textcolor{gray!80}{ClusComp} & \textcolor{gray!80}{2.01} & \textcolor{gray!80}{70.1} & \textcolor{gray!80}{63.3} & \textcolor{gray!80}{31.9} & \textcolor{gray!80}{44.4} & \textcolor{gray!80}{58.4} & \textcolor{gray!80}{53.6} \\
    \rowcolor{green!8}
    \textbf{ClusComp$^+$} & 2.01 & \textbf{74.8} & \textbf{66.7} & \textbf{34.8} & \textbf{49.6} & \textbf{63.4} & \textbf{57.8} \\ 
    \midrule
    \textbf{Llama-3-70B} & 16.00 & 82.5 & 86.7 & 60.4 & 66.3 & 80.9 & 75.4 \\
    QuIP & 2.00 & 65.3 & 48.9 & 26.5 & 40.9 & \textbf{61.7} & 48.7 \\
    PB-LLM & 1.70 & 56.5 & 49.9 & 25.8 & 34.9 & 53.1 & 44.1 \\
    BiLLM & 1.10 & 58.2 & 46.4 & 25.1 & 37.5 & 53.6 & 44.2 \\
    \rowcolor{green!8}
    \textcolor{gray!80}{ClusComp} & \textcolor{gray!80}{1.14} & \textcolor{gray!80}{56.9} & \textcolor{gray!80}{32.5} & \textcolor{gray!80}{20.6} & \textcolor{gray!80}{32.3} & \textcolor{gray!80}{51.6} & \textcolor{gray!80}{39.7} \\
    \rowcolor{green!8}
    \textbf{ClusComp$^+$} & 1.14 & \textbf{69.5} & \textbf{57.2} & \textbf{30.1} & \textbf{44.2} & 56.0 & \textbf{51.4} \\
    \bottomrule
    \end{tabular}
    \caption{Zero-shot accuracy on ultra-low-bit LLMs.}
    \label{tab: limit detailed number}
    \end{center}
\end{table*}

\begin{table*}[t]
    \centering
    \footnotesize
    \begin{tabular}{llcc}
    \toprule
    \multirow{2}{*}{\textbf{Method}} & \multirow{2}{*}{\textbf{\#Bit}} & \textbf{WikiText2}  & \textbf{GSM8K} \\
    & & PPL $\downarrow$ & Acc $\uparrow$ \\
    \midrule
    LoRA & 16.00 & 5.08 & 36.9 \\
    \cmidrule(lr){1-4}
    QLoRA & 4.25 + 0.40 & 5.70 & 35.1 \\
    LoftQ & 4.25 + 0.40 & \textbf{5.24} & 35.0 \\
    \rowcolor{green!8}
    \textbf{ClusComp} & \textbf{4.15} & 5.26 & \textbf{41.0} \\
    \cmidrule(lr){1-4}
    QLoRA & 3.25 + 0.40 & 5.73 & 32.1 \\
    LoftQ & 3.25 + 0.40 & 5.63 & 32.9 \\
    \rowcolor{green!8}
    \textbf{ClusComp} & \textbf{3.38} & \textbf{5.37} & \textbf{39.9} \\
    \cmidrule(lr){1-4}
    QLoRA & 2.25 + 0.40 & NAN & NAN \\
    LoftQ & 2.25 + 0.40 & 7.85 & 20.9 \\
    \rowcolor{green!8}
    \textbf{ClusComp} & 2.54 & \textbf{5.78} & \textbf{37.2} \\
    \rowcolor{green!8}
    \textbf{ClusComp} & \textbf{2.29} & 6.10 & 36.0 \\
    \bottomrule
    \end{tabular}
    \caption{In-domain finetuning performance on Llama-2-7B. Two bits are shown for baselines since the LoRA modules aren't merged to the quantized LLMs. The first and second numbers denote the quantized LLM and the converted bits from LoRA modules.}
    \label{tab: wiki gsm8k}
\end{table*}

\begin{table*}[t]
    \centering
    \begin{threeparttable}
    \footnotesize
    \begin{tabular}{lr|rrrrr|rrrrr}
    \toprule
    \multirow{3}{*}{\textbf{Method}} & \multirow{3}{*}{\textbf{\#Bit}} & \multicolumn{5}{c|}{\textbf{MMLU} (0-shot, Acc $\uparrow$)} &  \multicolumn{5}{c}{\textbf{MMLU} (5-shot, Acc $\uparrow$)} \\
     \cmidrule(lr){3-7} \cmidrule(lr){8-12}
    & & \textbf{Hums.} & \textbf{STEM} & \textbf{Social} & \textbf{Other} & \textbf{Avg} & \textbf{Hums.} & \textbf{STEM} & \textbf{Social} & \textbf{Other} & \textbf{Avg}\\
    \midrule
     \textbf{Llama-1-7B} & 16.00 & 32.4 & 26.6 & 31.4 & 37.2 & 32.1 & 33.3 & 29.8 & 37.8 & 38.0 & 34.6 \\
     \cmidrule(lr){1-12}
     GPTQ-LoRA & 4.50 & 35.7 & 30.9 & 38.0 & 44.0 & 37.1 & 33.8 & 31.3 & 37.4 & 42.2 & 36.0 \\
     QA-LoRA & 4.50 & \textbf{36.9} & \textbf{31.4} & \textbf{40.3} & \textbf{44.9} & \textbf{38.3} & 36.6 & 32.4 & \textbf{44.8} & \textbf{44.9} & \textbf{39.4} \\
     PEQA & \textbf{4.00} & - & - & - & - & - & 34.9 & 28.9 & 37.5 & 40.1 & 34.8 \\
     \rowcolor{green!8}
     \textbf{ClusComp} & 4.15 & \textbf{36.9} & \textbf{31.4} & 40.0 & 44.2 & 38.0 & \textbf{36.8} & \textbf{34.1} & 42.7 & 43.9 & 39.1 \\
     \cmidrule(lr){1-12}
     GPTQ-LoRA & 3.50 & 31.5 & 28.9 & 31.8 & 36.8 & 32.2 & 31.6 & 30.1 & 35.6 & 39.8 & 34.0 \\
     QA-LoRA & 3.50 & 36.0 & \textbf{34.1} & \textbf{42.0} & 42.3 & 38.3 & 35.6 & 30.5 & \textbf{41.5} & 42.7 & 37.4 \\
     \rowcolor{green!8}
     \textbf{ClusComp} & \textbf{3.38} & \textbf{38.2} & 32.7 & 41.2 & \textbf{45.4} & \textbf{39.2} & \textbf{36.3} & \textbf{31.4} & 41.3 & \textbf{43.0} & \textbf{37.8} \\
     \cmidrule(lr){1-12}
     GPTQ-LoRA & 2.50 & 24.1 & 22.1 & 22.5 & 23.7 & 23.2 & 23.4 & 26.2 & 26.4 & 28.4 & 25.8 \\
     QA-LoRA & 2.50 & 26.4 & 25.5 & 25.6 & 28.7 & 26.5 & 27.3 & 26.1 & 26.1 & 30.3 & 27.5  \\
     \rowcolor{green!8}
     \textbf{ClusComp} & \textbf{2.29} & \textbf{32.6} & \textbf{29.7} & \textbf{34.4} & \textbf{37.0} & \textbf{33.3} & \textbf{31.1} & \textbf{30.1} & \textbf{37.8} & \textbf{37.2} & \textbf{33.7} \\
     \midrule
     \textbf{Llama-2-7B} & 16.00 & 38.9 & 32.9 & 46.6 & 44.9 & 40.7 & 43.0 & 36.4 & 51.4 & 52.2 & 45.5  \\
     \cmidrule(lr){1-12}
     QA-LoRA & 4.50 & 41.1 & 35.4 & 50.2 & 50.1 & 43.9 & 42.1 & 34.4 & 49.1 & 50.3 & 43.9 \\
     \rowcolor{green!8}
     \textbf{ClusComp} & \textbf{4.15} & \textbf{41.6} & \textbf{36.3} & \textbf{52.3} & \textbf{51.1} & \textbf{44.9} & \textbf{42.8} & \textbf{38.1} & \textbf{52.2} & \textbf{53.1} & \textbf{46.1} \\
     \midrule
     \textbf{Llama-2-13B} & 16.00 & 48.1 & 42.7 & 60.5 & 59.5 & 52.3 & 53.3 & 44.1 & 63.3 & 61.0 & 55.3 \\
     \cmidrule(lr){1-12}
     QA-LoRA & 4.50 & 48.2 & 41.7 & 60.4 & 58.7 & 51.9 & 48.0 & 43.0 & 59.7 & 57.4 & 51.7  \\
     \rowcolor{green!8}
     \textbf{ClusComp} &  \textbf{4.09} &  \textbf{49.2} &  \textbf{42.9} &  \textbf{61.6} &  \textbf{60.2} &  \textbf{52.9} &  \textbf{52.4} &  \textbf{43.2} &  \textbf{62.9} &  \textbf{61.6} &  \textbf{54.7}  \\
    \bottomrule
    \end{tabular}
    \caption{General-domain finetuning performance, with baseline results from \citet{qalora}.}
    \label{tab: mmlu}
    \end{threeparttable}
\end{table*}

\begin{table*}[t]
    \begin{center}
    \scriptsize
    \begin{tabular}{lrlll}
    \toprule
    \textbf{Method} & \textbf{\#Bit} & \textbf{Llama-2-7B} & \textbf{Llama-2-13B} & \textbf{Llama-2-70B} \\
    \midrule
    - & 16.00 & 5.47 & 4.88 & 3.31 \\
    \cmidrule(lr){1-5}
    SqueezeLLM & 4.27 & 5.57 & 4.96 & - \\
    \rowcolor{green!8}
    \textbf{ClusComp} & $\leq$ \textbf{4.14} & \textbf{5.54} (4.14) & \textbf{4.94} (4.09) & - \\
    \cmidrule(lr){1-5}
    SqueezeLLM & 3.02 & 6.18 & 5.36 & 3.77 \\
    \rowcolor{green!8}
    \textbf{ClusComp} & $\leq$ \textbf{2.89} & \textbf{5.86} (2.89) & \textbf{5.18} (2.81) & \textbf{3.72} (2.72) \\
    \cmidrule(lr){1-5}
    SqueezeLLM & 2.22 & 10.79 & 7.91 & 4.99 \\
    SqueezeLLM & 2.05 & 13.64 & 8.56 & 5.38 \\
    SqueezeLLM & 2.01 & 35.49 & 41.02 & 9.44 \\
    \rowcolor{green!8}
    \textbf{ClusComp} & $\leq$ 2.00 & \textbf{7.50} (2.00) & \textbf{6.17} (1.99) & \textbf{4.83} (1.85) \\
    \bottomrule
    \end{tabular}
    \end{center}
    \caption{The perplexity of Llama-2 on WikiText2. The values in the brackets are the exact bits of ClusComp for different LLMs. The SqueezeLLM results are taken from \citet{squeezellm}.}
    \label{tab: squeezellm}
\end{table*}

\begin{table*}[t]
    \begin{center}
    \scriptsize
    \begin{tabular}{lrcccc}
    \toprule
    \multirow{3}{*}{\textbf{Method}} &  \multirow{3}{*}{\textbf{\#Bit}} & \multicolumn{2}{c}{\textbf{Llama-2-7B}} & \multicolumn{2}{c}{\textbf{Llama-2-13B}} \\
    \cmidrule(lr){3-4} \cmidrule(lr){5-6}
    & & \textbf{MMLU} (5-shot $\uparrow$) & \textbf{CSR} (0-shot $\uparrow$) & \textbf{MMLU} (5-shot $\uparrow$) & \textbf{CSR} (0-shot $\uparrow$) \\
    \midrule
    - & 16.00 & 46.0 & 67.9 & 55.6 & 70.3 \\
    \cmidrule(lr){1-6}
    GPTQ-AdaDim & 4.13 & 45.3 & 67.7 & 54.6 & 70.1 \\
    \rowcolor{green!8}
    \textbf{ClusComp} & $\leq$ 4.14 & \textbf{45.6} & \textbf{68.2} & \textbf{55.1} & \textbf{70.8} \\
    \cmidrule(lr){1-6}
    GPTQ-AdaDim & 3.13 & 41.3 & 66.4 & \textbf{52.3} & 68.7 \\
    \rowcolor{green!8}
    \textbf{ClusComp} & $\leq$ \textbf{2.89} & \textbf{43.2} & \textbf{67.2} & \textbf{52.3} & \textbf{69.5} \\
    \bottomrule
    \end{tabular}
    \end{center}
    \caption{The accuracy of quantized LLMs on MMLU and four commonsense reasoning (CSR) tasks (PIQA, HellaSwag, WinoGrande and ARC-easy). Following AdaDim, we use lm-eval v0.3.0 \citep{lmeval} for the evaluation. The GPTQ-AdaDim results are taken from \citet{adadim}.}
    \label{tab: adadim}
\end{table*}

\begin{table*}[t]
    \begin{center}
    \scriptsize
    \begin{tabular}{lrrllll}
    \toprule
    \textbf{Method} & \textbf{Bit for codebook} & \textbf{Avg. Bit} & \textbf{2-7B} & \textbf{2-70B} & \textbf{3-8B} & \textbf{3-70B} \\
    \midrule
    - & -  & 16.00 & 5.47 & 3.31 & 6.12 & 2.90 \\
    \cmidrule(lr){1-7}
    Best baseline & -  & 4.13 & 5.58 & 3.39 & 6.50 & 3.30 \\
    \rowcolor{green!8}
    ClusComp & 16 & $\leq$ 4.14 & 5.54 (4.14) & 3.40 (4.03) & 6.39 (4.13) & 3.12 (4.03) \\
    \rowcolor{green!8}
    ClusComp & 8 & $\leq$ 4.11  & 5.54 (4.11) & 3.40 (4.03) & 6.39 (4.10) & 3.13 (4.03)\\
    \rowcolor{green!8}
    ClusComp & 4 & $\leq$ 4.07  & 5.59 (4.07) & 3.43 (4.02) & 6.52 (4.07) & 3.26 (4.02) \\
    \rowcolor{green!8}
    ClusComp & 2 & $\leq$ 4.05  & 25.44 (4.05) & 5.64 (4.01) & 1.2e5 (4.05) & 96.52 (4.01) \\
    \cmidrule(lr){1-7}
    Best baseline & - & $\leq$ 2.13 & 35.07 (2.00) & 4.64 (2.13) & 85.10 (2.00) & 11.68 (2.13) \\
    \rowcolor{green!8}
    ClusComp & 16 & $\leq$ 2.07 & 7.50 (2.00) & 4.37 (2.07) & 12.33 (2.01) & 7.61 (2.07) \\
    \rowcolor{green!8}
    ClusComp & 8 & $\leq$ 2.04  & 7.50 (1.92) & 4.37 (2.04) & 12.33 (1.92) & 7.63 (2.04) \\
    \rowcolor{green!8}
    ClusComp & 4 & $\leq$ 2.03  & 7.63 (1.86) & 4.42 (2.03) & 12.77 (1.86) & 7.64 (2.03) \\
    \rowcolor{green!8}
    ClusComp & 2 & $\leq$ 2.02  & 6.5e3 (1.83) & 21.07 (2.02) & 1.7e5 (1.83) & 2.3e4 (2.02) \\
    \bottomrule
    \end{tabular}
    \end{center}
    \caption{The perplexity of ClusComp with quantized codebook on WikiText2. The results of the best baseline are taken from Table \ref{tab: wiki ppl}. The values in the brackets are the exact bits for different LLMs. We can observe: (1) 8-bit codebook offers the same perplexity as 16-bit's; (2) 4-bit codebook slightly hurts the performance, but is still comparable to the best baseline at the 4-bit level and outperforms the best baseline at the 2-bit level. (3) The results of the 2-bit codebook are not acceptable.}
    \label{tab: codebook quantization}
\end{table*}

\begin{table*}[t]
    \begin{center}
    \footnotesize
    \begin{tabular}{llrccc}
    \toprule
    \textbf{Method} & \textbf{Setting} & \textbf{Bit} & \textbf{WikiText2} $\downarrow$ & \textbf{C4} $\downarrow$ & \textbf{PTB} $\downarrow$ \\
    \midrule
    - & - & 16.00 & 9.5 & 14.8 & 16.3 \\
    \midrule
    GPTQ & w4g128 & 4.13 & \textbf{9.5} & 14.8 & 17.1 \\
    AWQ & w4g128 & 4.13 & 9.9 & 15.3 & \textbf{16.9} \\
    \rowcolor{green!8}
    \textcolor{gray!80}{\textbf{ClusComp}$^{-}$} & \textcolor{gray!80}{s4n65500} & \textcolor{gray!80}{4.13} & \textcolor{gray!80}{9.9} & \textcolor{gray!80}{\textbf{13.6}} & \textcolor{gray!80}{17.6} \\
    \rowcolor{green!8}
    \textbf{ClusComp} & s4n65500 & 4.13 & 9.7 & \textbf{13.6} & 17.7 \\
    \midrule
    GPTQ & w3g128 & 3.13 & 13.0 & 19.5 & 28.4 \\
    AWQ & w3g128 & 3.13 & 11.7 & 17.9 & \textbf{20.2} \\
    \rowcolor{green!8}
    \textcolor{gray!80}{\textbf{ClusComp}$^{-}$} & \textcolor{gray!80}{s6n65500} & \textcolor{gray!80}{2.87} & \textcolor{gray!80}{14.3} & \textcolor{gray!80}{16.2} & \textcolor{gray!80}{31.9} \\
    \rowcolor{green!8}
    \textbf{ClusComp} & s6n65500 & \textbf{2.87} & \textbf{10.7} & \textbf{15.3} & 22.0 \\
    \midrule
    GPTQ & w2g128 & 2.13 & 83.7 & 3.1e3 & 2.0e2 \\
    AWQ & w2g128 & 2.13 & 1.6e6 & 2.0e6 & 2.2e6 \\
    \rowcolor{green!8}
    \textcolor{gray!80}{\textbf{ClusComp}$^{-}$} & \textcolor{gray!80}{s8n35000} & \textcolor{gray!80}{2.14} & \textcolor{gray!80}{7.7e2} & \textcolor{gray!80}{6.1e3} & \textcolor{gray!80}{9.2e2} \\
    \rowcolor{green!8}
    \textbf{ClusComp} & s8n35000 & 2.14 & \textbf{14.6} & \textbf{21.8} & \textbf{27.5} \\
    \bottomrule
    \end{tabular}
    \end{center}
    \caption{The perplexity of the Llama-3-8B backbone in LLaVA-Next-8B, with baseline results from \citet{llama3quant}.}
    \label{tab: multimodal ppl}
\end{table*}

\section{Experimental Details}
\label{sec: experimental details}
\subsection{Clustering}
We use the K-means implementation from the Faiss library \citep{faiss}. The number of iterations is set to 20, with all default settings for other arguments.

\subsection{Block-wise error minimization}
For all LLMs, 128 calibration sentences with a length of 2048 tokens are randomly selected from the WikiText-2 training set \citep{wikitext}. The detailed hyper-parameters are listed in Table \ref{tab: hyperparameters for block-wise}. Only the codebooks are trained while keeping all other parameters (from the embedding layer, output layer and normalization layers) frozen.
\begin{table*}[t]
  \centering
  \small
  \begin{tabular}{lcc}
  \toprule
  \textbf{Hyper-parameter} & \textbf{Block-wise error minimization} & \textbf{Recovery training} \\
  \midrule
  Optimizer & \multicolumn{2}{c}{AdamW \citep{adamw, adam}} \\
  Weight decay & \{\underline{0}, 0.1, 0.01\} & 0 \\
  LR & \{1e-5, 5e-5, \underline{1e-4}, 5e-4\} & 1e-5 \\
  LR scheduler & constant & cosine \\
  Warmup ratio & 0 & 0 \\
  Max grad norm & - & 0.3 \\
  Sequence length & 2048 & 4096 \\
  Number of samples & 128 & 8192 \\
  Epochs & 20 & 1 \\
  Batch size & 8 & 8 \\
  \bottomrule
  \end{tabular}
  \caption{Hyper-parameters used for the block-wise error minimization and recovery training steps. The underlined \underline{settings} generally perform well for different scales of LLMs.}
  \label{tab: hyperparameters for block-wise}
  \centering
\end{table*}

\subsection{Recovery training}
\label{subsec: recovery}
For the recovery training, we randomly sample 1024 sentences with a length of 4096 tokens from a subset of RedPajama \citep{redpajama}\footnote{\href{https://huggingface.co/datasets/togethercomputer/RedPajama-Data-1T-Sample}{RedPajama data.}}. Then we train the compressed LLMs to predict the next token by only tuning the codebook parameters. The hyperparameters used in this step are listed in Table \ref{tab: hyperparameters for block-wise}.

\subsection{In-domain finetuning}
\label{subsec: in-domain}
We follow the settings from \citep{loftq}, and finetune the compressed LLM on the training set of WikiText2 and on the training set of GSM8K. The hyperparameters for finetuning are listed in Table \ref{tab: hyperparameters for wiki gsm8k}. We evaluate the finetuned model on the validation set of WikiText2 and on the test set of GSM8K every epoch and report the best perplexity or accuracy.

\begin{table*}[t]
  \centering
  \small
  \begin{tabular}{l|cc|cc}
  \toprule
  \textbf{Hyper-parameter} & \textbf{WikiText-2} & \textbf{GSM8K} & \textbf{Alpaca-GPT3.5} & \textbf{Alpaca-GPT4} \\
  \midrule
  Optimizer & \multicolumn{2}{c|}{AdamW} & \multicolumn{2}{c}{AdamW} \\
  Weight decay & \multicolumn{2}{c|}{0.1} & \multicolumn{2}{c}{0} \\
  LR & \multicolumn{2}{c|}{\{0.7, \underline{1}, 3\}$\times 10^{-4}$} & \multicolumn{2}{c}{\{2, 4, \underline{6}, 8\}$\times 10^{-5}$} \\
  LR scheduler & \multicolumn{2}{c|}{cosine} & \multicolumn{2}{c}{cosine} \\
  Warmup ratio & \multicolumn{2}{c|}{3\%} & \multicolumn{2}{c}{6\%} \\
  Epochs or max steps & 3 epochs & 6 epochs & 10K steps & 2 epochs \\
  Batch size & 64 & 16 & \multicolumn{2}{c}{16} \\
  Max sequence length & 1024 & 512 & \multicolumn{2}{c}{2048} \\
  \bottomrule
  \end{tabular}
  \caption{Hyperparameters for the finetuning on Llama-2-7B. The underlined \underline{settings} generally performs well for different bit levels. We report the average performance from three random runs in this paper.}
  \label{tab: hyperparameters for wiki gsm8k}
  \centering
\end{table*}

\subsection{General-domain finetuning}
\label{subsec: general-domain}
The finetuning hyper-parameters are listed in Table \ref{tab: hyperparameters for wiki gsm8k}, which is similar to the ones in QA-LoRA \citep{qalora} on Alpaca-GPT3.5, or to the ones in LISA \citep{lisa} on Alpaca-GPT4.

\begin{figure*}[t]
  \centering
  \includegraphics[width=0.98\textwidth]{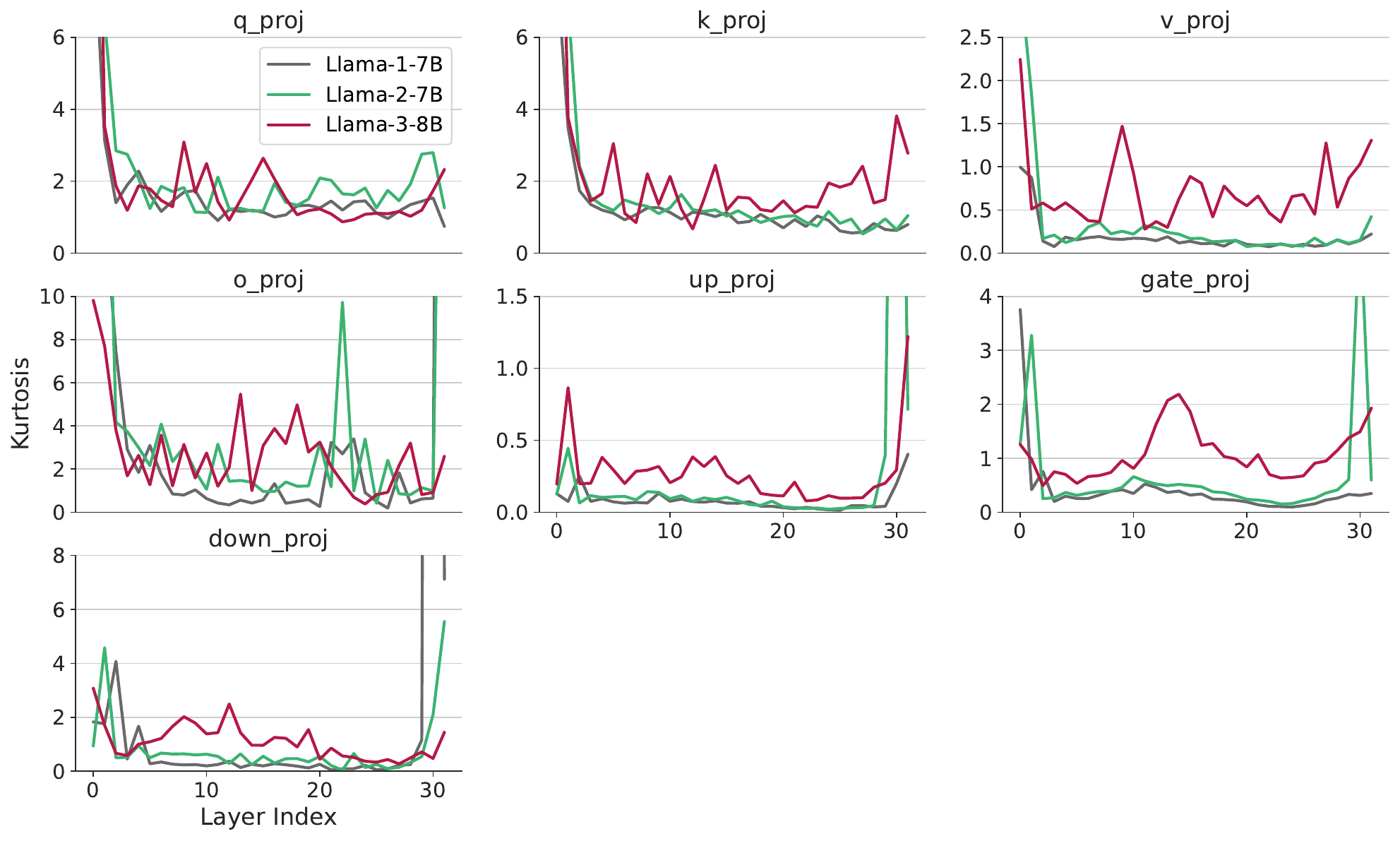}
  \caption[]{The kurtosis across various layers in different Llama series reveals three key observations: (1) Layers at either the beginning or the end of LLMs tend to exhibit higher kurtosis values; (2) In the majority of layers, the kurtosis follows a consistent trend across Llama series, with Llama-3 showing the highest values, followed by Llama-2, and then Llama-1; (3) Different types of layers display varying scales of kurtosis, suggesting that a bit allocation strategy that accounts for quantization difficulty could yield better results. We leave the exploration of this idea to future work.}
  \label{fig: layer kurtosis}
\end{figure*}

\begin{figure*}[t]
  \centering
  \includegraphics[width=0.98\textwidth]{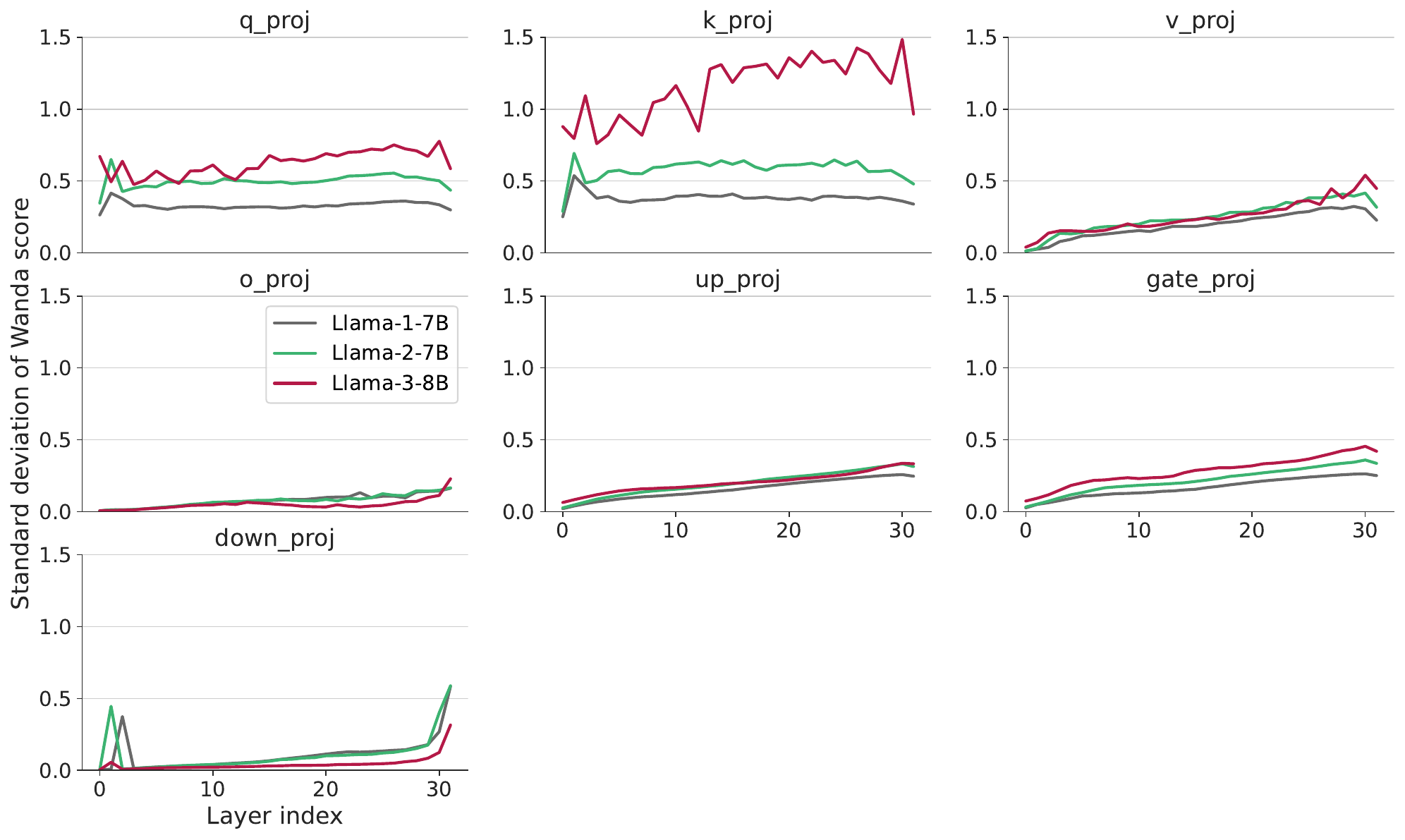}
  \caption[]{The standard deviation of Wanda score across various layers in different Llama series reveals three key observations: (1) Deeper layers tend to exhibit higher standard deviation; (2) Three layers (query, key and gate) show a clear trend across Llama series, with Llama-3 showing the highest standard deviation, followed by Llama-2, and then Llama-1; (3) The other four layers show a similar standard deviation for all Llama series.}
  \label{fig: wanda}
\end{figure*}

\clearpage
\onecolumn
\begin{lstlisting}[language=Python, caption={PyTorch code for the linear layer of ClusComp. All data type is 16-bit.}, label={lst: linear}]
class ClusCompLinear(nn.Module):
    def __init__(self, in_features, out_features, num_clusters, cluster_dim, bias):
        super().__init__()
        self.out_features = out_features
        self.in_features = in_features
        self.deficiency = out_features % cluster_dim # If the out_features is not dividable by cluster_dim
        if self.deficiency > 0:
            self.deficiency = cluster_dim - self.deficiency

        num_codes = in_features * (out_features + self.deficiency) // cluster_dim
        self.codebook = nn.Parameter(torch.empty((num_clusters, cluster_dim), dtype=torch.bfloat16) # trainable
        code = torch.empty((num_codes,), dtype=torch.uint16)
        self.register_buffer('code', code) # non-trainable
        if bias:
            self.bias = nn.Parameter(torch.empty(out_features))
        else:
            self.register_parameter('bias', None)

    def forward(self, x):
        vectors = self.codebook[self.code]
        if self.deficiency > 0:
            weight = vectors.view(self.in_features, -1)[:, :-self.deficiency]
        else:
            weight = vectors.view(self.in_features, -1)
            
        if self.bias is not None:
            out = torch.matmul(x, weight) + self.bias
        else:
            out = torch.matmul(x, weight)
        return out
\end{lstlisting}

\end{document}